\pdfoutput=1

\documentclass[11pt]{article}

\usepackage{acl}

\usepackage{times}
\usepackage{latexsym}

\usepackage[T1]{fontenc}
\usepackage{graphicx}
\usepackage{amsmath}
\usepackage{amssymb}
\usepackage{booktabs}
\usepackage{multirow}
\usepackage{adjustbox}
\usepackage{soul}
\usepackage[table]{xcolor}
\usepackage{changepage,threeparttable}
\usepackage{booktabs}
\usepackage{bbding}
\usepackage{pifont}
\usepackage{microtype}
\usepackage{stfloats}
\usepackage{graphicx}
\usepackage{xspace}

\usepackage{hyperref}\hypersetup{linkcolor=red,anchorcolor=blue,citecolor=blue}

\usepackage[utf8]{inputenc}
\usepackage{xcolor}
\usepackage{etoolbox}

\newcommand{\method}{{\em SnapNTell}\xspace} 

\newcommand{\cmark}{\ding{51}}%
\newcommand{\xmark}{\ding{55}}%

\usepackage{microtype}

\title{\method: Enhancing Entity-Centric Visual Question Answering with \\
Retrieval Augmented Multimodal LLM}

\author{Jielin Qiu$^{1,2}$\thanks{Work done while at Meta}, ~Andrea Madotto$^{1}$, ~ Zhaojiang Lin$^{1}$, ~Paul A. Crook$^{1}$, ~Yifan Ethan Xu$^{1}$,\\
\textbf{~Xin Luna Dong$^{1}$,  ~Christos Faloutsos$^{2}$, ~Lei Li$^{2}$, ~Babak Damavandi$^{1}$, ~Seungwhan Moon$^{1}$ }\\
  $^{1}$ Meta Reality Labs \& FAIR, Meta
  ~$^{2}$Carnegie Mellon University
  \\
  \small{$\left\{\text{jielinq,leili,christos}\right\}$@cs.cmu.edu, $\left\{\text{andreamad8,zhaojiang,pacrook,ethanxu,lunadong,shanemoon}\right\}$@meta.com}
  }

\begin{document}
\maketitle
\begin{abstract}
Vision-extended LLMs have made significant strides in Visual Question Answering (VQA). Despite these advancements, VLLMs still encounter substantial difficulties in handling queries involving long-tail entities, with a tendency to produce erroneous or hallucinated responses. In this work, we introduce a novel evaluative benchmark named \textbf{SnapNTell}, specifically tailored for entity-centric VQA.  This task aims to test the models' capabilities in identifying entities and providing detailed, entity-specific knowledge. We have developed the \textbf{SnapNTell Dataset}, distinct from traditional VQA datasets: (1) It encompasses a wide range of categorized entities, each represented by images and explicitly named in the answers; (2) It features QA pairs that require extensive knowledge for accurate responses. The dataset is organized into 22 major categories, containing 7,568 unique entities in total. For each entity, we curated 10 illustrative images and crafted 10 knowledge-intensive QA pairs.  To address this novel task, we devised a scalable, efficient, and transparent retrieval-augmented multimodal LLM. Our approach markedly outperforms existing methods on the SnapNTell dataset, achieving a 66.5\% improvement in the BELURT score. We will soon make the dataset and the source code publicly accessible.
\end{abstract}

\vspace{-5pt}
\section{Introduction}

\begin{figure}[t]
  \centering
  \includegraphics[width=0.9\linewidth]{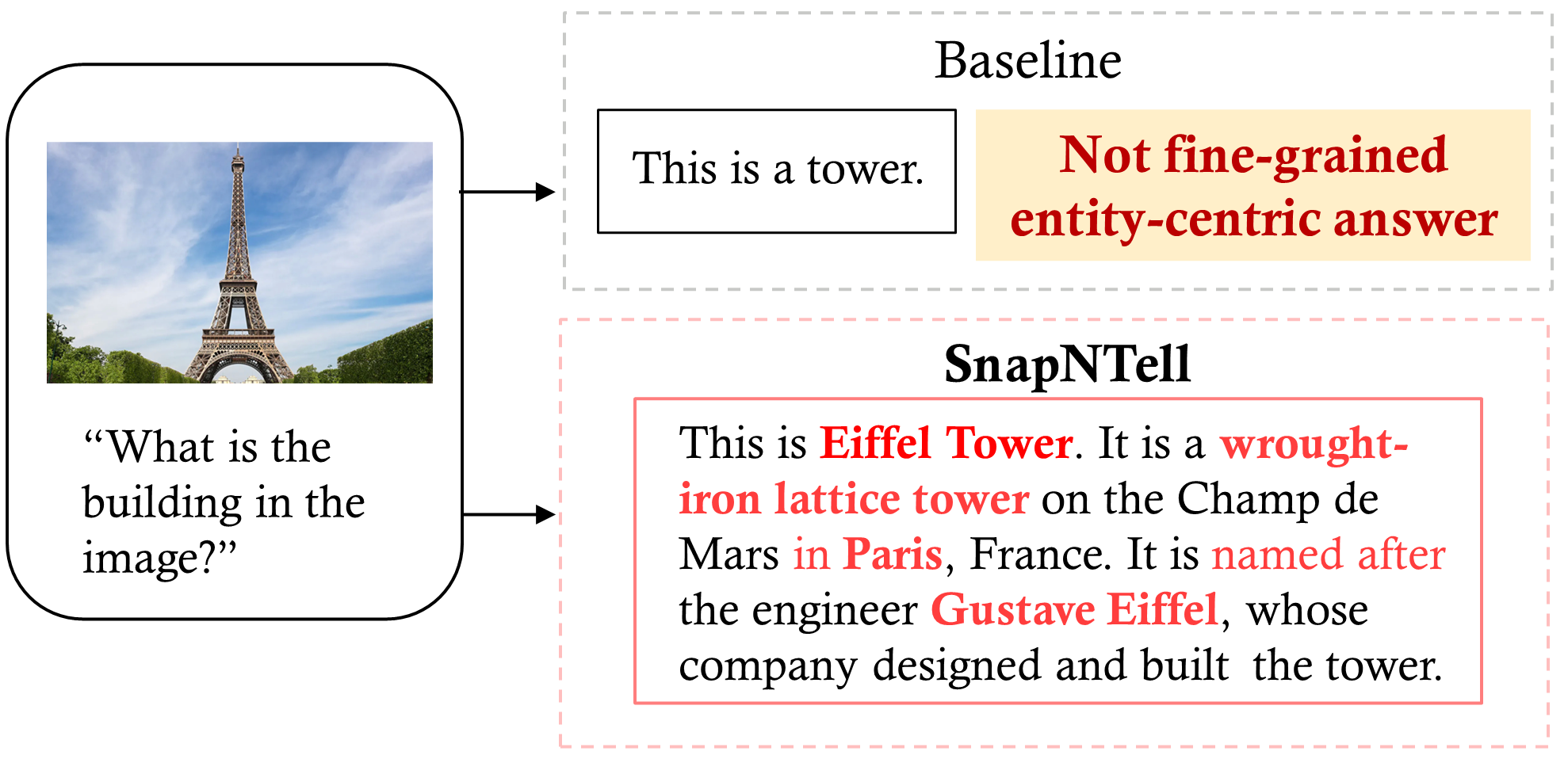}
  \caption{Comparing SnapNTell with existing methods reveals a distinctive focus. In the SnapNTell benchmark, the answers are predominantly \textbf{entity-centric}, characterized by a greater depth of knowledgeable information pertaining to the specific entity depicted in the image as the answer.}
  \label{Fig:intro}
  \vspace{-18pt}
\end{figure}

Vision-extended LLMs have shown significant advancements, excelling at capturing complex semantics and context-aware attributes needed for intricate tasks. However, their abilities in factual VQA tasks, which demand accurate, concrete answers about real-world entities and phenomena, expose certain limitations. Particularly, torso-to-tail or long-tail entities, which constitute a large proportion of real-world data but appear infrequently in training datasets, pose a challenge. This scarcity in representation often leads to VLLMs resorting to generating plausible but incorrect or imaginative content in their outputs, a problem that manifests as ``hallucinations" within the context of model responses.
To ensure the confident deployment of VLLMs in practical scenarios, there is an urgent need for dedicated research that not only recognizes but actively strives to tackle and reduce instances of hallucinations, especially in the context of factual queries involving these long-tail entities.

The lack of publicly available evaluation datasets specifically tailored to assess models' ability in recognizing real-world long-tailed entities presents a notable gap in VQA. Existing datasets fall short in serving this purpose due to a narrow range of entity categories, the prevalence of overly simplistic yes/no QA pairs, and a general lack of entity specificity, often using broad terms like ``Tiger" instead of more specific ones like ``Siberian Tiger". To address this gap, we introduce a novel evaluation task called \textbf{SnapNTell}, which focuses on entity-centric knowledge-based VQA. The SnapNTell benchmark has been designed to evaluate models' abilities in accurately identifying entities and generating responses that showcase a deep understanding of these entities. To support this task, we have curated a new evaluation dataset that departs from existing datasets in two crucial ways:
(1) It includes a wide range of fine-grained and categorized entities, each accompanied by corresponding images and clear mention of the entity name within the answer sets. (2) It features QA pairs designed to prompt knowledge-intensive responses, moving beyond the binary yes/no format to challenge and assess the depth of the model's comprehension.

Furthermore, the limitations identified in factual query generation underscore the need for new solutions to address the problem of hallucinations. Recent advancements suggest that retrieval-based approaches hold significant promise in this regard \citep{Guu2020REALMRL,Srinivasan2022QUILLQI,Yang2023InferenceWR,Yang2023ReViLMRV}. These methods enhance LLMs by integrating external knowledge sources or incorporating retrieval mechanisms to access relevant information from extensive knowledge bases. The synergy between the advanced inference capabilities of LLMs and the wealth of external knowledge has the potential to significantly reduce issues related to long-tail entities and, consequently, decrease the occurrence of hallucinatory responses.

In this work, we aim to propose an evaluation task to investigate the model's ability to recognize real-world long-tailed entities and provide knowledge-intensive answers. We also propose a retrieval-augmented method to reduce hallucinations and enhance the precision and trustworthiness of generated responses.

Our contribution is summarized as follows:
\vspace{-5pt}
\newcommand{\scalable}{{\bf Scalable}:}
\newcommand{\effective}{ {\bf Effective}:}
\newcommand{\explainable} { {\bf Explainable}:}
\begin{itemize}
    \item \textbf{SnapNTell task.} We propose a novel task for \underline{\textbf{entity-centric}} VQA, specifically designed to assess the proficiency of models in accurately identifying and generating responses that exhibit a deep comprehension of these identified entities.
    \vspace{-5pt}
    \item \textbf{SnapNTell model.} We proposed a retrieval-augmented multimodal LLM, devised as a baseline model capable of undertaking the SnapNTell task, which is scalable, effective, and explainable.
    \vspace{-5pt}
    \item \textbf{SnapNTell dataset.} We collected a new evaluation dataset with distinctive characteristics, which stands out for two key features: (1) It encompasses a diverse range of fine-grained entities, each accompanied by corresponding representative images. (2) The question-answer pairs contain knowledge-intensive responses with entity names specifically mentioned in the answer sets.
    \vspace{-10pt}
    \item Our model demonstrates superior performance on the SnapNTell dataset, surpassing current methodologies with a 66.5\% improvement in BELURT score.
\end{itemize}

\section{Related Works}\label{sec:related-work}

\paragraph{Knowledge-based VQA}

Research in vision-language tasks, which necessitate understanding image content to answer questions, has seen significant advancements over recent years. Beginning with datasets like FVQA \citep{Wang2016FVQAFV}, which extracted facts from pre-established knowledge bases, the field has progressed to more challenging ones like the OK-VQA dataset \citep{Marino2019OKVQAAV}, encompassing diverse knowledge categories. MultiModalQA  \citep{Talmor2021MultiModalQACQ} introduced complexity with questions demanding cross-modal reasoning over snippets, tables, and images. The successor of OK-VQA, AOK-VQA \citep{Schwenk2022AOKVQAAB}, raises the bar by providing questions that transcend simple knowledge base queries. ManyModalQA \citep{Hannan2020ManyModalQAMD} shifts the focus to answer modality selection, MIMOQA \citep{Singh2021MIMOQAMI} emphasizes multimodal answer extraction, and WebQA \citep{Chang2021WebQAMA} introduces real-world knowledge-seeking questions, albeit with some limitations regarding entity categorization and granularity. More comparison details can be found in Section~\ref{sec:compare_datasets}.

\vspace{-5pt}
\paragraph{Multimodal LLMs}

Integrating visual understanding into text-based LLM typically combines them with a visual encoder and uses image captioning datasets for alignment \citep{Koh2023GroundingLM,Wu2023NExTGPTAM,Chowdhery2022PaLMSL}. Techniques like adapter-based tuning \citep{Alayrac2022FlamingoAV} and prefix tuning \citep{Tsimpoukelli2021MultimodalFL} allow these models to process visual inputs while maintaining their linguistic capabilities, without requiring full model retraining \citep{Yin2023ASO}.

\vspace{-5pt}
\paragraph{Retrieval-augmented LLM} 
Previous studies have explored retrieval augmentation in text-only settings or image captioning tasks. \citet{Guu2020REALMRL} introduced a retriever for language models to access large corpus during various stages. \citet{Srinivasan2022QUILLQI} showed retrieval-augmented queries enhance LLMs' context understanding. \citet{Yasunaga2023RetrievalAugmentedML} and \citet{Yang2023InferenceWR} developed methods for integrating multimodal documents and speeding up LLM inference, respectively. \citet{Yang2023ReViLMRV} created a visual language model, inspired by Flamingo \citep{Alayrac2022FlamingoAV}, for image captioning with external database retrieval. Similarly, \citet{Gui2021KATAK} combined implicit and explicit knowledge in an encoder-decoder setup to improve answer generation.

\vspace{-5pt}
\paragraph{Open-domain visual entity recognition}   

\citet{Hu2023OpendomainVE} developed OVEN for associating images with Wikipedia entities via text queries, while \citet{Chen2023CanPV} introduced INFOSEEK, a dataset for Visual Question Answering focused on informational queries. While OVEN is proficient in entity recognition using a knowledge base, INFOSEEK mainly supplies factual responses. Our study seeks to merge these strengths, creating detailed paragraphs that provide context for a more comprehensive understanding beyond basic facts.
More related work can be found in Appendix~\ref{sec:related-work-appendix}.

\section{SnapNTell Dataset}

\subsection{Entity Categorization}

To tackle the challenge of the new SnapNTell task, the first step involves creating a comprehensive dataset that represents a wide array of real-world entities. Our dataset creation methodology entails selecting a diverse set of entity names from various categories that mirror the diversity of the real world. This selection encompasses both commonly encountered entities and less frequently encountered ones.
We have identified 22 categories that adequately represent a cross-section of entities one might encounter in daily life. 
These categories include \textit{landmark, painting, sculpture, food, fruit, vegetable, mammal, amphibian, insect, fish, bird, reptile, celebrity, instrument, plant, electronics, tool, transportation, sport, book, household, and car}.
More details about the categories can be referred to Table~\ref{table:dataset-statistics} in the Appendix.

To populate each category with specific entities, we leveraged Wikipedia as a primary resource due to its extensive and detailed entries. (See Appendix~\ref{sec:appendix-more-dataset} for more details.) Our selection criteria are heavily biased towards specificity; for instance, in the category of mammals, we deliberately opted for precise names such as ``German Shepherd” or ``Alaskan Malamute” instead of the generic ``Dog”. This level of specificity is critical as it enables the model to demonstrate its capacity for fine-grained recognition and its ability to generate detailed, accurate information about each entity.
This dataset-building approach is what distinguishes our dataset from existing VQA datasets, which often lack fine-grained entities and specificity.

\subsection{Image collection}

The dataset comprises 22 primary categories, encapsulating a total of 7,568 unique entities. For each individual entity, a set of 10 images has been curated, where the statistic of the entity list is shown in Table~\ref{table:dataset-statistics} in the Appendix.

\vspace{-5pt}
\paragraph{Filtering} 
Initially, a comprehensive list of entities, encompassing 22 primary categories, was compiled, in a total of 14,910 diverse entities.
Then the entity list underwent filtering by cross-referencing each entry with its corresponding Wikipedia page. Entities lacking valid Wikipedia pages were subsequently removed from the list.
For each corresponding entity, images were sourced from Creative Commons (CC).
Further filtering was conducted by removing entities that didn't have a sufficient number of images obtained via Google Image Search engine.
The collected metadata was stored in a CSV file containing essential information such as image URLs, source page URLs, renamed image names, and the corresponding Wikipedia page URLs. After filtering, the final number of entities in the SnapNTell dataset is 7,568. (More filtering details can be found in Appendix~\ref{sec:appendix-more-filtering}.)

\subsection{Knowledge-intensive Question-Answer Pairs}

In our SnapNTell dataset, we considered five types of questions:
\vspace{-5pt}
\begin{itemize}
    \item \textbf{Static facts (absolute facts, discrete facts).} These are objective facts that are concrete and are not contingent on other conditions. They can usually be answered with a unique answer. i.e., ``When was he (Barack Obama) born?"
    \vspace{-5pt}
    \item \textbf{Narrative facts.}  These facts encompass comprehension of larger contexts (e.g., song lyrics, movie plot). They are factual in the sense that the content of the narrative should accurately reflect the source material or events, but a correct answer is usually not unique, as they can vary in their level of detail and focus. i.e., ``What is the plot of that (`The Godfather')?"
    \vspace{-5pt}
    \item \textbf{Dynamic facts.}  These are facts that are subject to change over time. i.e., ``What is the Yelp customer rating of it (the Eleven Madison Park restaurant) in NYC?"
    \vspace{-5pt}
    \item \textbf{Procedural facts.}  These are usually answers to ``how” questions, outlining a sequence of steps to accomplish a task. While the steps may not be unique and could be subjective, the answer can still be classified as logical or nonsensical. Note that these facts may sometimes overlap with dynamic facts or narrative facts, i.e., ``How do you check the battery level of my item (Ray-Ban Stories Glasses)?"
    \vspace{-5pt}
    \item \textbf{Subjective facts. (opinion-based facts).}  These “facts” are not objective indisputable facts, but based on individual perspectives or experience. Recommendations fall in this category. While there’s generally no single correct answer to questions seeking subjective facts, it still requires the system to understand the topic and provide reasonable answers grounded by world facts. i.e., ``Why do you like it (Niagara Falls)?"
\end{itemize}

To construct a comprehensive and knowledge-intensive QA dataset, we employ a three-step process. Firstly, we extracted and condensed pertinent information from Wikipedia for each entity, i.e., the summary of the introduction, the caption of the image, etc. (See Appendix~\ref{sec:appendix-more-dataset} for more details). 
Following similar approaches proposed by LLaVA \citep{Liu2023VisualIT}, \citet{Dettmers2023QLoRAEF} is utilized to generate QA pairs for each entity automatically based on five pre-defined question types, ensuring diversity and informativeness.
Then, we enlisted three annotators (2 male and 1 female) from Amazon SageMaker to assess QA pair quality and make necessary revisions to meet specific criteria. 
The responsibilities of these annotators include: (1) ensuring that the images and QA pairs are semantically aligned, (2) validating the accuracy of the provided answers, (3) making sure the questions are free of particular entity names but demanding such specificity in the answers, (4) assessing if the modified QA pairs adhere to the criteria for knowledge-intensive content, and (5) removing specific entity-related details from the questions. This last step guarantees that \underline{the question queries cannot be answered without} \underline{understanding the accompanying visual context}.

\paragraph{Quality and consistency} 
In order to verify the quality of the QA pairs, we conducted a quality evaluation by randomly choosing 1,000 QA pairs from our dataset. We assigned three independent human evaluators (1 male, 2 female) from Amazon SageMaker to review these pairs for accuracy [\textit{accurate, inaccurate}] and agreement on
whether to save the QA pair by Fleiss’ Kappa \citep{Fleiss1971MeasuringNS}. The outcome of this assessment revealed 98\% accuracy and $\kappa = 0.95$ agreement rate among the evaluators, demonstrating a significant degree of uniformity in the quality of the QA pairs.

\subsection{Statistics and Analysis of Our Dataset}\label{sec:dataset-tatistics}

\paragraph{Entity statistics}

To provide a clear summary of this comprehensive dataset, we have condensed the details of the entity list into Table~\ref{table:dataset-statistics} and Figure~\ref{Fig:Statistics_catergory} (in Appendix~\ref{sec:appendix-more-statistics}). Our analysis indicates that the dataset displays a well-balanced distribution across different categories, enhancing its balanced and diverse characteristics. Such a balanced and diverse composition enhances the representativeness of our proposed evaluation dataset.

\paragraph{Popularity}

The importance of entity popularity in search engines is a key aspect to consider, similar to examining the head, torso, and tail sections of knowledge bases within search engine frameworks. As demonstrated in Figure~\ref{Fig:ave_pageview} in Appendix~\ref{sec:appendix-more-statistics}, we use the average Wikipedia pageviews per entity over the last 60 days as the metric. This average is calculated by summing up the pageviews and then dividing by the number of entities. The insights from Figure~\ref{Fig:ave_pageview} reveal that entities in the celebrity category have the highest average popularity.
For a broader comparison among different categories, we also present a comprehensive analysis of total pageviews for all categories in Figure~\ref{Fig:statistics_pageview} in Appendix~\ref{sec:appendix-more-statistics}, which shows that the celebrity category remains at the forefront in terms of overall entity popularity. This is attributed to the combination of a higher number of entities in this category and the generally higher popularity of each entity within it.

\begin{table*}[t]\small
\caption{More detailed comparison with existing knowledge-based VQA datasets. \textit{Anonymity} means whether the question already contains a knowledge clue related to the entity in question. (* Unclear)}
\vspace{-5pt}
\centering
\begin{adjustbox}{width=0.99\linewidth}
\begin{tabular}{l|c|c|c|c|c|c|c}
\toprule
Dataset & Categories & Unique Entity & QA Pairs  & Images & Average Ans Length & Number of Images / Entity & Anonymity\\
\midrule
ViQuAE  & 3 & 2,400 & 3,700  & 3,300 &1.8  &*  &\xmark \\
Encyclopedic VQA (test)     &12 &*  &5,750 &5,750  &3.2 &*  &  \xmark  \\
\cellcolor[HTML]{FBBE78}SnapNTell (Ours) & \cellcolor[HTML]{FBBE78} 22 & \cellcolor[HTML]{FBBE78} 7,568 &  \cellcolor[HTML]{FBBE78} 75,680  & \cellcolor[HTML]{FBBE78} 75,680 & \cellcolor[HTML]{FBBE78}25.7   & \cellcolor[HTML]{FBBE78}10  & \cellcolor[HTML]{FBBE78}\cmark\\
\bottomrule
\end{tabular}
\label{table:compare_KVQA}
\end{adjustbox}
\vspace{-5pt}
\end{table*}

\begin{table}[t]\small
\caption{Comparison with existing VQA datasets \textit{Knowledge} means the QA pairs are knowledgeable, not simple yes/no answers or selection questions. \textit{Entities} means whether there are fine-grained entities specifically contained in answers. \textit{Categorization} means the entities are categorized, not randomly crawled online.}
\vspace{-5pt}
\centering
\begin{adjustbox}{width=0.99\linewidth}
\begin{tabular}{l|c|c|c}
\toprule
Dataset & Knowledge & Entities & Categorization  \\
\midrule
VQA 2.0 \citep{Goyal2016MakingTV} &  &   &  \\
GQA \citep{Hudson2019GQAAN} & &   &   \\
OK-VQA \citep{Marino2019OKVQAAV} &  &   &   \\
ManyModalQA \citep{Hannan2020ManyModalQAMD}  & \cmark  &  &   \\
MultiModalQA \citep{Talmor2021MultiModalQACQ} & \cmark &   &   \\
MIMOQA \citep{Singh2021MIMOQAMI} & \cmark  &  &   \\
A-OKVQA \citep{Schwenk2022AOKVQAAB} & \cmark &  &     \\
WebQA \citep{Chang2021WebQAMA} & \cmark & \cmark  & \cmark  \\
ViQuAE \citep{Lerner2022ViQuAEAD} & \cmark & \cmark  &  \cmark  \\
Encyclopedic VQA \citep{Mensink2023EncyclopedicVV}     & \cmark & \cmark  &  \cmark  \\
\cellcolor[HTML]{FBBE78}SnapNTell (Ours) & \cellcolor[HTML]{FBBE78}\cmark &  \cellcolor[HTML]{FBBE78}\cmark & \cellcolor[HTML]{FBBE78}\cmark  \\
\bottomrule
\end{tabular}
\label{table:compare_VQAdataset}
\end{adjustbox}
\vspace{-10pt}
\end{table}

\subsection{Comparison with Existing VQA Datasets}\label{sec:compare_datasets}

\begin{figure}[t]
  \centering
  \includegraphics[width=0.9\linewidth]{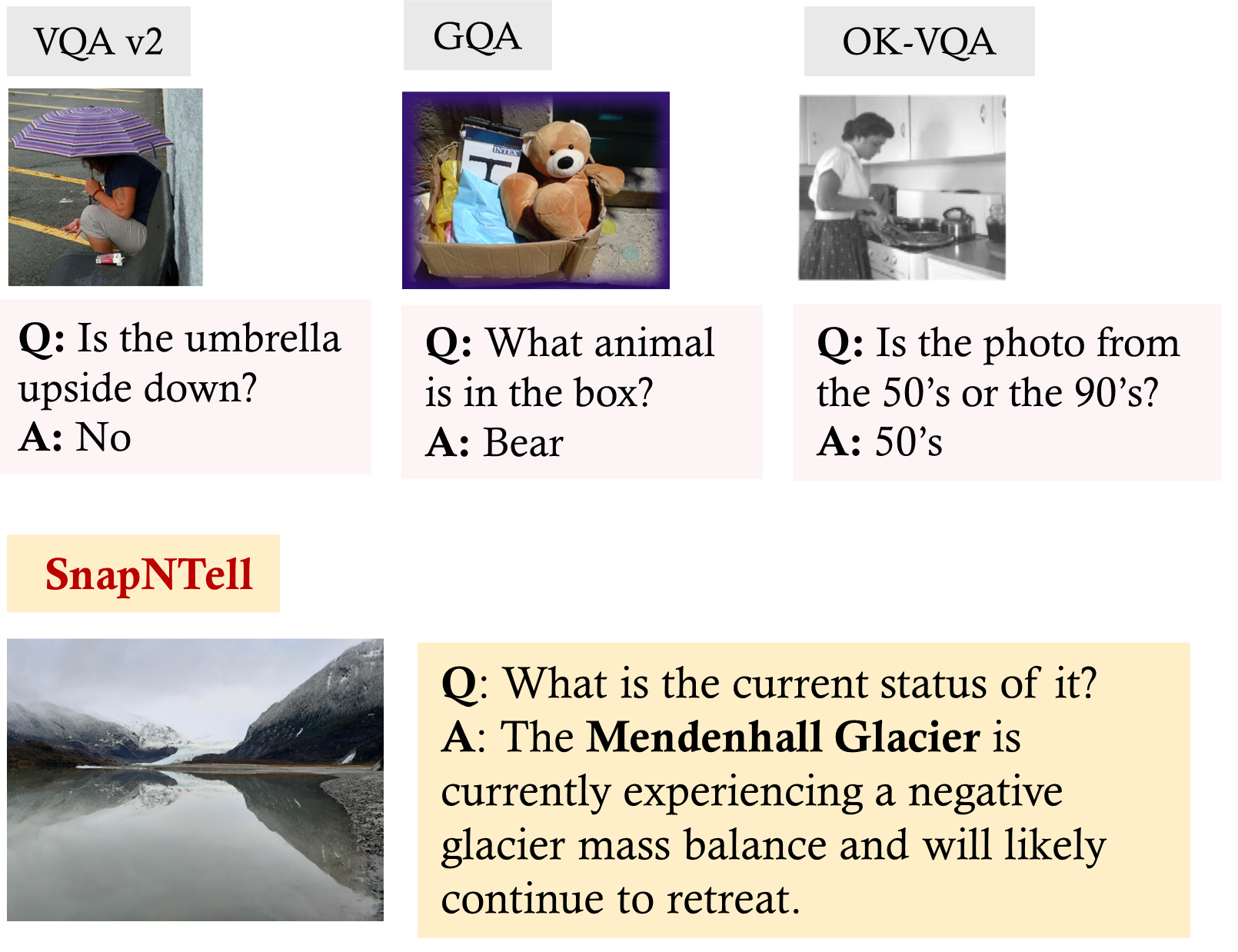}
  \caption{Comparison with existing datasets, where previous VQA datasets mostly focus on freeform answers (such as yes/no for verification questions and choice for selection questions).}
  \label{Fig:compare_dataset_example}
  \vspace{-15pt}
\end{figure}

In Table~\ref{table:compare_VQAdataset} and Figure~\ref{Fig:compare_dataset_example}, we present a comparison with existing VQA datasets. It is evident that some existing VQA datasets lack categorization, fine-grained entities, and knowledge-intensive answers, as observed in VQA 2.0 \citep{Goyal2016MakingTV} and GQA \citep{Hudson2019GQAAN}. OK-VQA \citep{Marino2019OKVQAAV} contains images that may not be sufficient to answer the questions, encouraging reliance on external knowledge resources. However, the answers in OK-VQA are often simplistic binary (yes/no) responses or selections from the questions. A-OKVQA \citep{Schwenk2022AOKVQAAB}, the successor of OK-VQA, aims to provide questions that require commonsense reasoning about the depicted scene but use general object names in the answers. MultiModalQA \citep{Talmor2021MultiModalQACQ} focuses on cross-modal knowledge extraction but relies on question templates for question generation. ManyModalQA \citep{Hannan2020ManyModalQAMD} focuses on answer modality choice rather than knowledge aggregation or extraction. In MIMOQA \citep{Singh2021MIMOQAMI}, the task of extracting a multimodal answer is not necessarily knowledge-intensive. WebQA \citep{Chang2021WebQAMA} does have categorization but lacks fine-grained entities in many QA pairs, resulting in more general questions and answers. Our proposed SnapNTell differs by including a wide range of fine-grained entities with representative images and explicit entity names in the answer sets. Additionally, it incorporates question-answer pairs that demand knowledge-intensive responses, going beyond simplistic binary answers.
Examples of our dataset can be found in Figure~\ref{Fig:our_example} in Appendix~\ref{sec:appendix-more-statistics}.

ViQuAE \citep{Lerner2022ViQuAEAD} and Encyclopedic VQA \citep{Mensink2023EncyclopedicVV} both incorporate entity-level knowledge-based information along with categorization. Therefore, we performed a more in-depth analysis comparing them in Table~\ref{table:compare_KVQA}. Our dataset surpasses these in terms of the variety of categories, the number of distinct entities, and the overall number of QA pairs. Additionally, our dataset boasts a higher count of images and a longer average length for answers. Specifically, our dataset is structured to include 10 images for each entity, whereas the exact number of images per entity in ViQuAE and Encyclopedic VQA remains unspecified. Most notably, our dataset's questions are highly anonymous, implying that they do not reveal any knowledge hints about the entity. This design ensures that the questions cannot be straightforwardly answered without interpreting the image data, setting our dataset apart from both ViQuAE and Encyclopedic VQA.

\section{Method}\label{sec:method}

In this section, we will introduce the details of our proposed retrieval-augmented multimodal LLM model. The architecture of our model is shown in Figure~\ref{Fig:propsoed_approach-main} (larger figure in Appendix~\ref{sec:method-appendix} due to space limit). Our model can be considered twofold: (1) \textbf{Retrieval augmentation}. Given the input image-question pair, we retrieve useful entity-centric information within knowledge sources. (2) \textbf{Entity-centric knowledge-based answer generation}. The retrieved information will be combined with the image and question together to generate a knowledgeable answer.

\begin{figure}[tp]
  \centering
  \includegraphics[width=0.99\linewidth]{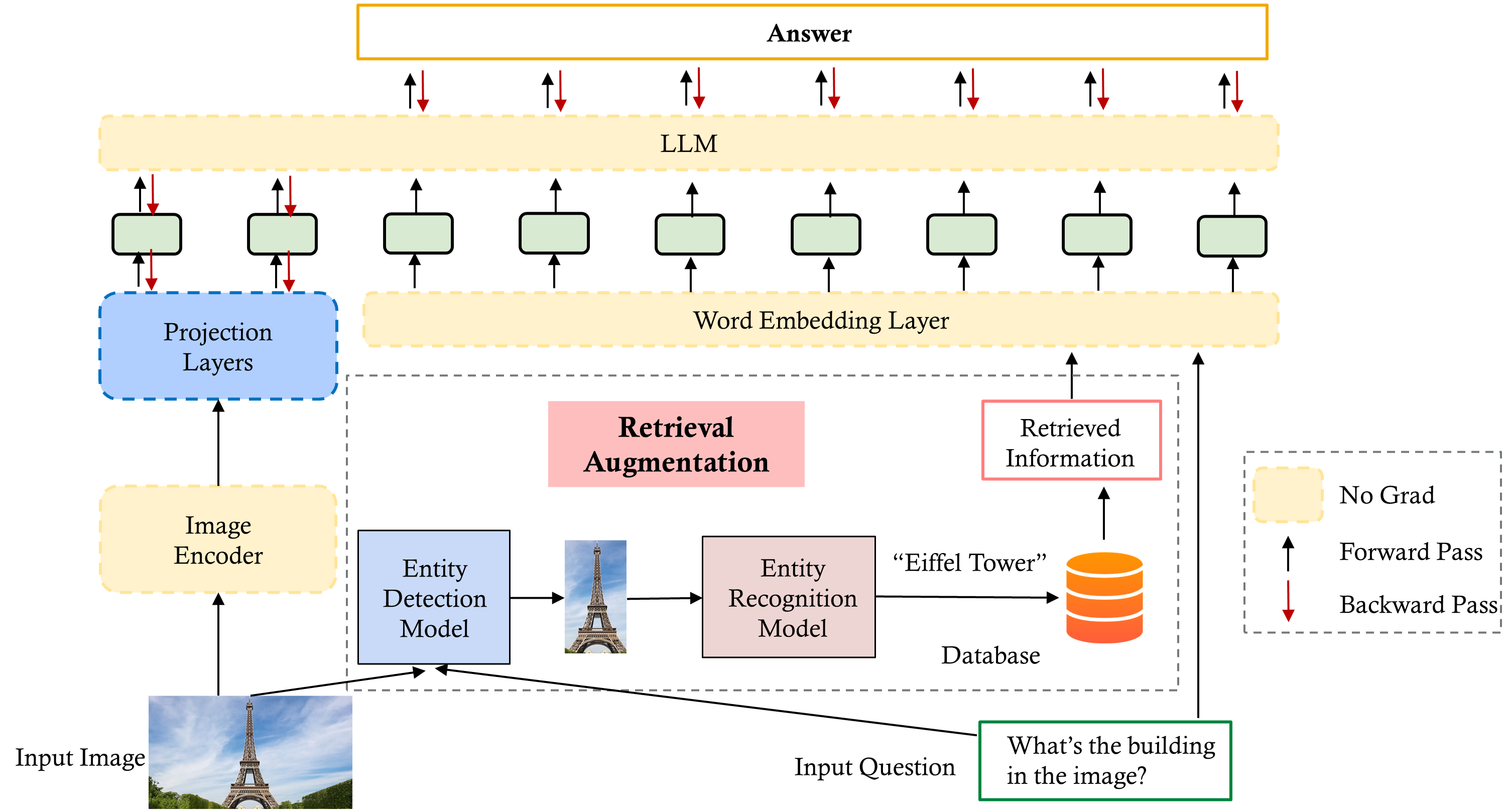}
  \caption{Our SnapNTell model architecture takes an image-question pair as input. It begins with retrieval augmentation to source relevant information about the entity in the image. This information, along with the question, feeds into the word embedding layer. Text embeddings merge with image-projected embeddings before entering the LLM, culminating in a knowledgeable answer as the output.}
  \label{Fig:propsoed_approach-main}
  \vspace{-15pt}
\end{figure}

\subsection{Retrieval Augmentation}

The retrieval augmentation process can be subdivided into: (i) Semantic region extraction via language-guided object detection, (ii) Entity recognition via image retrieval, and (iii) Knowledge retrieval via multi-source aggregation.

\vspace{-5pt}
\paragraph{Semantic Region Extraction via Language-Guided Object Detection}
To improve recognition performance, we focus on extracting specific image regions containing the entity, rather than general image-level recognition. We employ a language-guided object detection model, i.e., GLIP \citep{Li2021GroundedLP}, for language-guided object detection, extracting regions relevant to textual queries by understanding the query context. This targeted approach ensures precise region extraction, enhancing the system's accuracy and contextual relevance.

\vspace{-5pt}
\paragraph{Entity Recognition via Image Retrieval}

We construct a similarity index using CLIP embeddings \citep{Radford2021LearningTV} and Faiss \citep{Johnson2017BillionScaleSS} for indexing. Our database, built on the WIT dataset \citep{Srinivasan2021WITWI}, maps CLIP image embeddings to their text descriptions, leveraging Faiss's robust similarity search capabilities.
After setting up the indexing database, given an input query image $I$, we perform a $k$-nearest neighbor retrieval based on cosine similarity. The retrieval outcomes are represented as $\mathcal{R}(I)=\{\left(i_1, c_1\right), \cdots,\left(i_k, c_k\right)\}$, where for each $j$ within the range of $1$ to $k$, $i_j$ and $c_j$ correspond to the retrieved image and its associated caption, respectively.
By comparing $I$ with similar images from the database, we identify the entity in the image region, which enables precise image-level entity recognition.

\vspace{-5pt}
\paragraph{Knowledge Retrieval via Multi-Source Aggregation}

Facing diverse user queries, we gather extra information to compile resources for accurate responses. Some queries require up-to-date information, not present in existing databases. We then turn to external sources to collect critical data like ``year built," ``description," and more. By using Knowledge Graph (KG) and web searches, we access relevant knowledge links, enriching our understanding of the specified image region, and improving our ability to comprehend and contextualize the extracted content.
More details of the method can be found in Appendix~\ref{sec:method-appendix}.

\subsection{Entity-centric Knowledge-based Answer Generation}

Following information collection, we enter the integration phase, blending the input image, question, and retrieved data to generate a knowledgeable response, which is illustrated in Figure~\ref{Fig:propsoed_approach-main}.
Our method enhances multimodal understanding by pre-training a LLM with image-text paired data. Taking cues from \citet{Moon2023AnyMALAE}, we employ lightweight adapters for each modality, converting inputs into the text token embedding space of the chosen LLM.

In our method, the LLM's text token embedding space morphs into a unified space, representing both text and image content, with each modality assigned 64 to 256 token embeddings. We freeze the LLM's parameters during alignment training to quicken convergence and retain the LLM's reasoning skills for inference. To ensure feature alignment, we use an image encoder, $g(\cdot)$, previously synchronized with a text embedding space, like in CLIP \citep{Radford2021LearningTV, Schuhmann2022LAION5BAO}. For text-image pairs $(\mathbf{X}_{\textit{text}}, \mathbf{X}_{\textit{image}})$, we align them using specific objectives and a projection module, like the Perceiver Resampler \citep{Alayrac2022FlamingoAV}, applied to the vision encoder as:
\vspace{-5pt}
\begin{equation}\small
\begin{aligned}
p(\mathbf{X}_{\textit{text}} | \mathbf{X}_{\textit{image}}) = \prod_{i=1}^{L} p_{\theta}(\mathbf{X}^{[i]}_{\textit{text}} | \mathbf{Z}_{\textit{image}}, \mathbf{Z}^{[1:i\small{-}1]}_{\textit{text}}) 
\end{aligned}
\end{equation}
\vspace{-5pt}
\begin{equation}\small
\begin{aligned}
\mathbf{Z}_{\textit{image}} = \textit{Proj}_{\theta}(h_{\textit{latents}}, g(\mathbf{X}_{\textit{image}}))
\end{aligned}
\end{equation}

\section{Experiments and Results}

\subsection{Experimental Setup} 

\paragraph{Evaluation Metrics} 

(1) In our evaluation process, the quality of the answers is first assessed using established NLP metrics such as BLEU \citep{Papineni2002BleuAM}, METEOR \citep{Denkowski2014MeteorUL}, ROUGE \citep{Lin2004ROUGEAP}, and BLEURT \citep{Sellam2020BLEURTLR,Pu2021LearningCM}. 
(2) Additionally, we incorporate accuracy and hallucination rate metrics from \citep{Sun2023HeadtoTailHK}. These metrics used \underline{GPT4} to automatically measure the proportion of questions for which the model provides correct answers or incorrect/partially incorrect answers, respectively. 
(3) We conduct human evaluation following \citet{ye2023mplugowl2,Moon2023AnyMALAE}.

\vspace{-5pt}
\paragraph{Model Setting} 
We chose LLaMA2 (70B) \citep{Touvron2023Llama2O} as our LLM. For image encoding, the CLIP image encoder (ViT-B/32) is employed \citep{Radford2021LearningTV,Schuhmann2022LAION5BAO}. Additional configurations comprise a batch size of 2,048, the integration of two resampler layers, and the use of 64 modality tokens.

\vspace{-5pt}
\paragraph{Model Training}

We used a cleaned subset of the LAION-2B dataset, filtered using the CAT method \citep{Radenovic2023FilteringDA} and with any detectable faces blurred \citep{laionfiltering}.
Significant resources are essential to scale pre-training to 70 billion parameter models on a substantial dataset of over 200 million instances. Often, this necessitates the utilization of an FSDP wrapper, as outlined in \citet{Dettmers2023QLoRAEF}, to distribute the model across multiple GPUs efficiently.
To optimize our training process, we employ quantization strategies, specifically 4-bit and 8-bit quantization techniques \citep{Dettmers2023QLoRAEF}, within our multimodal framework. In this approach, we maintain the LLM component of our model in a frozen state, allowing only the image modality tokenizers to be trainable. This strategy drastically reduces the memory requirements by an order of magnitude.
As a result of these optimizations, we can successfully train a 70 billion parameter model on a single GPU with 80GB VRAM, using a batch size of 4.

\subsection{Results and Discussion}

Table~\ref{table:results-main} displays the comparative results between the baseline models and our proposed method. Analysis of this table indicates that for every metric assessed, our retrieval-augmented multimodal LLM surpasses the performance of all existing baseline models. This strong performance emphasizes the efficiency of retrieval augmentation in producing responses enriched with entity-centric information, thereby illustrating its substantial impact on the task at hand.

\begin{table}[tp]
\centering
\caption{Performance comparison of different approaches on the SnapNTell dataset. }
\vspace{-5pt}
\begin{adjustbox}{width=0.99\linewidth}
\begin{tabular}{lrrrrr}\toprule
Method  &ROUGE $\uparrow$  &BLEU $\uparrow$ &METEOR $\uparrow$ &BLEURT $\uparrow$ \\
\midrule
Instruct-BLIP \cite{Dai2023InstructBLIPTG}  &10.72  &0.95 &7.59 &0.09 \\
BLIP2 \cite{Li2023BLIP2BL}  &15.00&0.52 &8.49 &0.16 \\
Mini-GPT4 \cite{Zhu2023MiniGPT4EV}  &26.12 &5.62 &25.55 &0.27 \\
LLaVA \cite{Liu2023VisualIT}  &26.86 &6.03 &26.97 &0.31 \\
Open-Flamingo \cite{Awadalla2023OpenFlamingoAO}  &30.57  &6.52 &22.53 &0.32 \\
COGVLM \cite{Wang2023CogVLMVE} & 30.25 &6.67  & 23.35  &0.31            \\
mPLUG-Owl2 \cite{ye2023mplugowl2} & 31.39  & 6.72  & 24.67   &0.33  \\
LLaVA 1.5 \cite{Liu2023ImprovedBW} &32.87  &6.94    &25.23  &0.33       \\
\cellcolor[HTML]{FBBE78}SnapNTell (ours)  &\cellcolor[HTML]{FBBE78}\textbf{35.28}  &\cellcolor[HTML]{FBBE78}\textbf{7.81} &\cellcolor[HTML]{FBBE78}\textbf{29.27} &\cellcolor[HTML]{FBBE78}\textbf{0.55} \\ 
\bottomrule
\end{tabular}
\end{adjustbox}
\label{table:results-main}
\vspace{-15pt}
\end{table}

Moreover, to gain deeper insights into which evaluation metric more accurately reflects the outcomes, we computed the Kendall correlation coefficient \citep{Kendall1938ANM,Knight1966ACM,Kendall1995KendallsAT}, comparing the results with those from the human evaluation in Section~\ref{sec:human_evaluation}. 
Kendall’s $\tau$ is a measure of the correspondence between two rankings. Values close to 1 indicate strong agreement, values close to -1 indicate strong disagreement.
Table~\ref{table:results-metric-effectiveness} revealed that both the ROUGE and BLEURT scores were more indicative in distinguishing the differences among various models. This finding suggests that these two metrics are particularly significant in evaluating model performance in a way that aligns closely with human judgment.

\begin{table}[tp]
\centering
\caption{Effectiveness of evaluation metrics.}
\vspace{-5pt}
\begin{adjustbox}{width=0.8\linewidth}
\begin{tabular}{lcccc}\toprule
  &ROUGE   &BLEU  &METEOR    &BELURT  \\
\midrule
$\tau$ &0.999 &0.799 &0.600 &0.999 \\
P\_value  &0.014 &0.050 &0.142 &0.014 \\ 
\bottomrule
\end{tabular}
\end{adjustbox}
\label{table:results-metric-effectiveness}
\vspace{-10pt}
\end{table}

\subsection{Ablation Study}

For a more in-depth understanding, we conducted several ablation studies to delve into the finer details of our approach.

\vspace{-5pt}
\paragraph{Effectiveness of Entity Detection} 
To assess the impact of entity detection (ED) in our model, we performed an ablation study. This involved comparing the performance of our approach with and without the ED component. As indicated in Table~\ref{table:results-ablation-ED}, our approach incorporating entity detection markedly surpasses the variant lacking this feature. This highlights the significant contribution and necessity of the entity detection step in our model's overall effectiveness.

\begin{table}[htp]
\centering
\caption{Ablation study on the effectiveness of entity detection (ED).}
\vspace{-5pt}
\begin{adjustbox}{width=0.9\linewidth}
\begin{tabular}{lrrrrr}\toprule
Method  &ROUGE $\uparrow$  &BLEU $\uparrow$ &METEOR $\uparrow$ &BELURT $\uparrow$ \\
\midrule
w/o ED &28.02 &3.73 &26.26 &0.45 \\
w/ ED  &\textbf{35.28}  &\textbf{7.81} &\textbf{29.27} &\textbf{0.55} \\ 
\bottomrule
\end{tabular}
\end{adjustbox}
\label{table:results-ablation-ED}
\vspace{-10pt}
\end{table}

\vspace{-5pt}
\paragraph{Head/Torso/Tail Entities}

Head knowledge pertains to well-established entities for which there is a wealth of available training data. Ideally, LLMs could be trained to possess this knowledge, facilitating efficient retrieval. On the other hand, torso-to-tail knowledge pertains to less-known or obscure entities, often characterized by scarce or non-existent training data. Providing access to such knowledge involves effectively determining when external information is necessary, retrieving the relevant knowledge efficiently, and seamlessly integrating it into responses.

To assess the performance improvement for head/torso/tail entities, we randomly selected 10\% entities for each category, where head/torso/tail entities are defined based on pageview statistics (popularity) in Section~\ref{sec:dataset-tatistics}. The results presented in Table~\ref{table:ablation_tail} clearly demonstrate that retrieval augmentation can significantly enhance performance across various entity types. Notably, the performance improvement for torso-to-tail entities far exceeds that of head entities, effectively addressing the challenge of hallucinations in long-tailed entities through retrieval augmentation.

\begin{table}[tp]\small
\caption{Ablation study on head/torso/tail entities, where RA is short for Retrieval Augmentation and $\Delta$ is the performance difference of with and without RA.}
\vspace{-5pt}
\centering
\begin{adjustbox}{width=0.9\linewidth}
\begin{tabular}{lrccc}\toprule
& &Accuracy $\uparrow$ &Hallucination $\downarrow$ \\
\toprule
\multirow{3}{*}{Head} 
&w/o RA &24.4 &75.6 \\
&w/ RA &27.1 &72.9 \\
&$\Delta$ (100\%) &\textcolor{orange}{11.1 \% $\uparrow$}  &\textcolor{cyan}{3.6 \% $\downarrow$} \\
\midrule
\multirow{3}{*}{Torso} 
&w/o RA &19.1 &80.9 \\
&w/ RA &22.7 &77.3 \\
&$\Delta$ (100\%) &\textcolor{orange}{18.8 \% $\uparrow$}  &\textcolor{cyan}{4.4 \% $\downarrow$} \\
\midrule
\multirow{3}{*}{Tail} 
&w/o RA &6.8 &93.2 \\
&w/ RA &12.6 &87.4 \\
&$\Delta$ (100\%) &\textcolor{orange}{85.3 \% $\uparrow$}  &\textcolor{cyan}{6.2 \% $\downarrow$ } \\
\bottomrule
\end{tabular}
\label{table:ablation_tail}
\end{adjustbox}
\vspace{-10pt}
\end{table}

\vspace{-5pt}
\paragraph{Performance of Different VQA Datasets }

To demonstrate the uniqueness of our SnapNTell dataset compared to existing VQA datasets, we analyzed the performance of various baseline models on both traditional VQA datasets and our SnapNTell dataset. 
According to the findings presented in Table~\ref{table:results-ablation-datasets}, the performance disparities among baseline models on existing datasets are not particularly marked. In contrast, on the SnapNTell dataset, we observed significantly larger differences and notably lower performance. This indicates that our SnapNTell dataset is particularly effective in evaluating the capabilities of different models to recognize entities and produce responses centered around these entities.
\vspace{-5pt}
\begin{table}[htp]
\centering
\caption{Ablation on the \underline{accuracy} performance of different VQA datasets.}
\vspace{-5pt}
\begin{adjustbox}{width=0.99\linewidth}
\begin{tabular}{lrrrrr}\toprule
Method  &VQAv2  &TextVQA  &OK-VQA &\cellcolor[HTML]{FBBE78}SnapNTell \\
\midrule
Instruct-BLIP \citep{Dai2023InstructBLIPTG} &-- &46.6 &55.5  &\cellcolor[HTML]{FBBE78}8.88 \\
BLIP2 \citep{Li2023BLIP2BL} &52.6 &43.1 &54.7  &\cellcolor[HTML]{FBBE78}16.16 \\
Flamingo \citep{Alayrac2022FlamingoAV} &56.3 &37.9 & 57.8  &\cellcolor[HTML]{FBBE78}32.17 \\
\bottomrule
\end{tabular}
\end{adjustbox}
\label{table:results-ablation-datasets}
\vspace{-10pt}
\end{table}

\subsection{Human Evaluation Results}\label{sec:human_evaluation}

In alignment with the methodology presented in \citet{ye2023mplugowl2,Moon2023AnyMALAE}, we involved a human evaluation process conducted by a panel of five human judges (3 male, 2 female). These judges were given specific instructions for their assessment, which encompassed three key aspects: (1) Recognition Accuracy, where they evaluated whether the model correctly identified the entity in the image relevant to the question; (2) Response Accuracy, in which they assessed the factual correctness of the model's responses while checking for any signs of hallucination \citep{Rawte2023ASO}; and (3) Pairwise Comparison, where judges selected the response that better addressed the given question in terms of contextual appropriateness and accuracy, categorizing responses as winning, tying, or losing.

\begin{figure}[tp]
  \centering
  \includegraphics[width=0.95\linewidth]{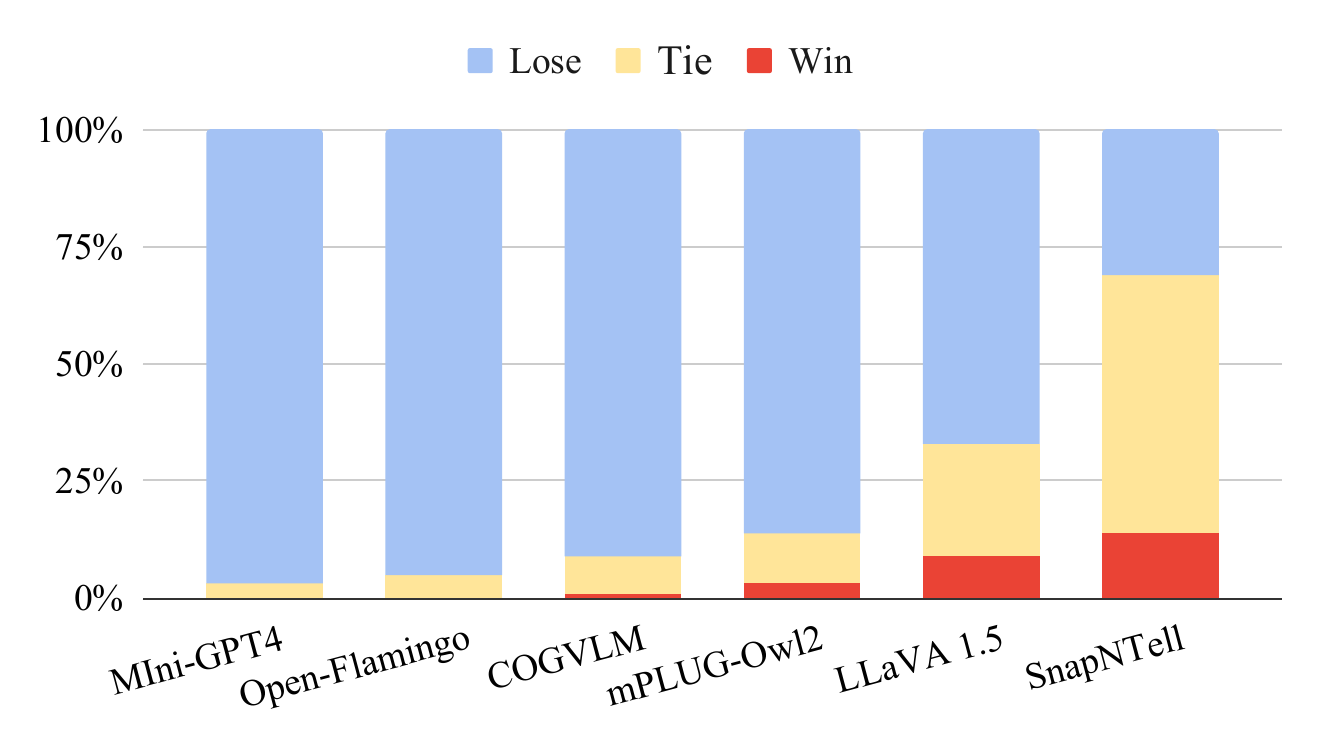}
  \caption{Human evaluation results on pairwise comparisons (\% win, tie, lose) with baseline outputs \textit{against} the manually annotated ground-truth from SnapNTell. }
  \label{Fig:human_eval}
  \vspace{-15pt}
\end{figure}

In our study, we conducted pairwise comparisons for each baseline model against ground-truth data across 1,000 samples. As depicted in Figure~\ref{Fig:human_eval}, our model outperforms the baselines by displaying a significantly smaller difference when measured against manually annotated ground-truth samples, highlighting its robustness.


\section{Conclusion}

In this work, we tackle the significant challenge VLLMs face with long-tail entity queries, which often lead to inaccurate or hallucinated responses. To address these issues, we introduce an entity-centric VQA task named SnapNTell. This task is designed to test models on entity recognition and their ability to provide detailed, entity-specific knowledge in their responses. We collected a unique evaluation dataset for this task, which distinguishes itself from existing VQA datasets by including a wide array of fine-grained categorized entities, supported by images and explicit entity mentions in the answers. This dataset emphasizes knowledge-intensive responses over simple binary answers. In addition, we propose a retrieval-augmented multimodal LLM solution for the SnapNTell task as an effective baseline. Our experimental results show that our model outperforms existing approaches, providing more accurate and coherent answers.

\clearpage
\section*{Limitations}
In this study, we introduce a novel SnapNTell task and its accompanying dataset, which features five unique types of questions, each paired with meticulously formulated answers. It's important to recognize that in cases involving human preferences, which are subjective by nature, the given answers might not represent the only correct options. Furthermore, the relevancy of some answers may diminish over time, highlighting the need for periodic updates to the dataset to ensure its ongoing relevance and accuracy. Our proposed method exhibited superior performance over existing baselines. However, human evaluation results suggest significant potential for further improvement. Although our approach often neared human-level performance, it did not consistently outperform human annotations, showing opportunities for future advancements.

\section*{Ethics Statement}

In this study, the dataset was sourced from publicly accessible databases, and all author details remain anonymous. We conscientiously excluded any content from our dataset that could be considered ethically sensitive or related to personal privacy, such as images depicting human faces.
To our understanding, and with careful consideration, we do not anticipate any detrimental applications arising from the findings or methodologies presented in this research.

\section*{Broader Impact}

Current models have made commendable progress in grasping the nuanced semantics and context-sensitive aspects of Visual Question Answering (VQA). However, their efficacy in factual VQA tasks, which require precise and factual answers about tangible entities and events, reveals certain deficiencies. This is especially true for torso-to-tail or long-tail entities. Despite their prevalence in the real world, these entities are underrepresented in training datasets, leading to a common issue where models produce plausible yet inaccurate or invented responses, a phenomenon often termed ``hallucinations" in the realm of model-generated content. Tackling and minimizing these hallucinations is vital for enhancing the trustworthiness and applicability of these models in practical scenarios.

The existing VQA datasets, however, are inadequate for evaluating a model's ability to recognize entities, as they do not explicitly highlight these entities within the dataset. Our newly introduced dataset bridges this gap. It is designed to test models' capabilities not just in identifying entities but also in generating informed and entity-aware responses. Furthermore, our proposed dataset might serve as resources for either pre-training or fine-tuning existing models, to improve their ability in recognizing entity-level real-world objects.

\bibliography{egbib}
\bibliographystyle{acl_natbib}

\newpage
\appendix
\onecolumn

\section{More Details about the Dataset Building}\label{sec:appendix-more-dataset}

\begin{figure*}[htp]
  \centering
  \includegraphics[width=0.99\linewidth]{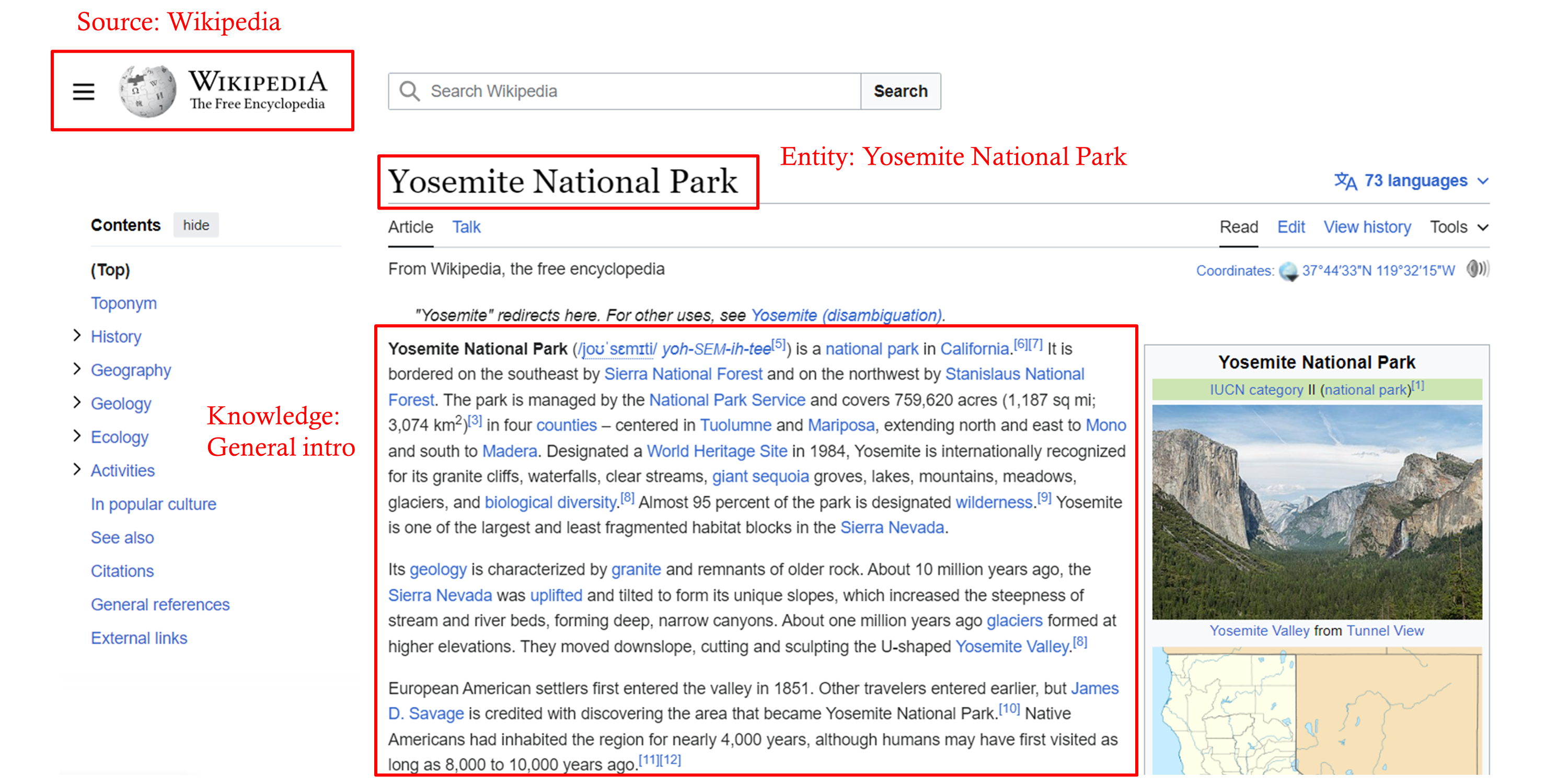}
  \includegraphics[width=0.99\linewidth]{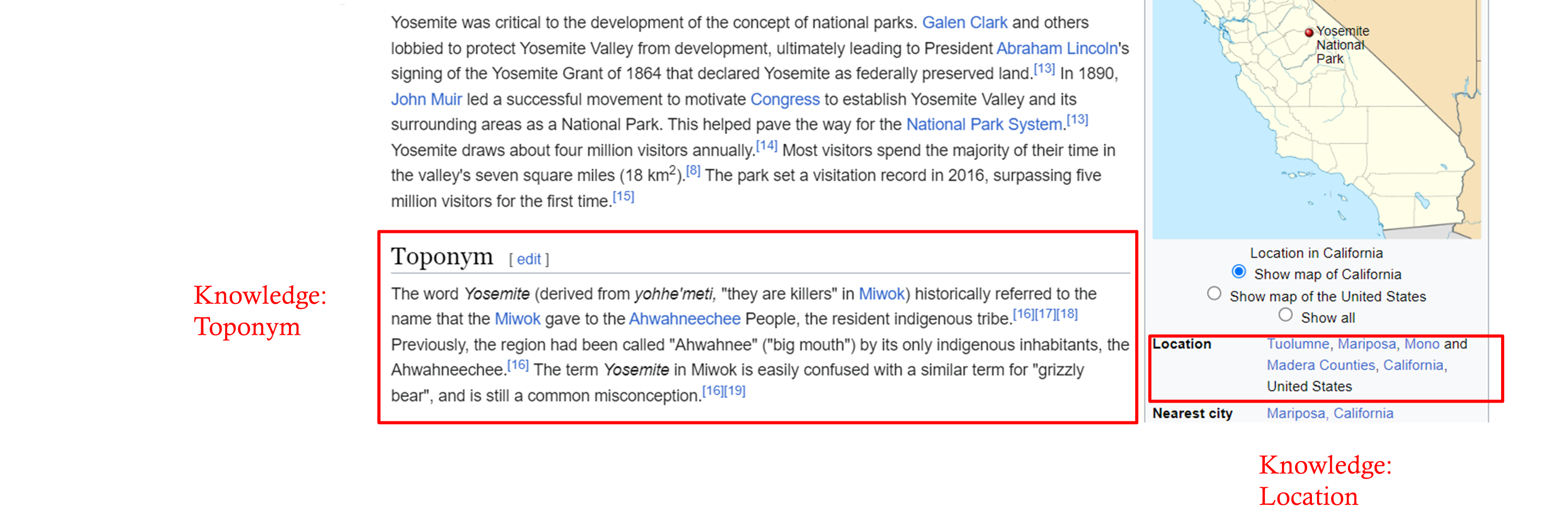}
  \caption{The pertinent information collected during dataset building, i.e., from Wikipedia for each entity, which includes the summary of the general introduction, toponym, lococation information, and so on.}
  \label{Fig:wiki}
  \vspace{-10pt}
\end{figure*}

More details about the dataset building process are shown in Figure~\ref{Fig:wiki}.

\section{More Details about the Filtering Process}\label{sec:appendix-more-filtering}

More details about the filtering process are shown in Table~\ref{table:more-filtering}.

\begin{table*}[htp]\centering
\caption{Filtering statistics of the entity dataset. [1st Wiki filtering]: removing ones without wiki page. [2nd Google filtering]: removing ones without enough images via google search API. [3rd Wiki filtering]: removing entity name with ambiguous wiki pages. }
\scriptsize
\begin{adjustbox}{width=0.9\linewidth}
\begin{tabular}{llrrrrr}
\toprule
&Main category &Original Entity &\cellcolor[HTML]{fff2cc}1st Wiki filtering  &\cellcolor[HTML]{d9ead3}2nd Google filtering &\cellcolor[HTML]{cfe2f3}3rd Wiki filtering \\
\midrule
\multirow{22}{*}{\textbf{Category}} &landmark &1595 &\cellcolor[HTML]{fff2cc}1000 &\cellcolor[HTML]{d9ead3}899 &\cellcolor[HTML]{cfe2f3}753 \\
&painting &1057 &\cellcolor[HTML]{fff2cc}367 &\cellcolor[HTML]{d9ead3}358 &\cellcolor[HTML]{cfe2f3}288 \\
&sculpture &300 &\cellcolor[HTML]{fff2cc}164 &\cellcolor[HTML]{d9ead3}164 &\cellcolor[HTML]{cfe2f3}134 \\
&food &883 &\cellcolor[HTML]{fff2cc}338 &\cellcolor[HTML]{d9ead3}337 &\cellcolor[HTML]{cfe2f3}271 \\
&fruit &361 &\cellcolor[HTML]{fff2cc}236 &\cellcolor[HTML]{d9ead3}233 &\cellcolor[HTML]{cfe2f3}180 \\
&vegetable &389 &\cellcolor[HTML]{fff2cc}290 &\cellcolor[HTML]{d9ead3}286 &\cellcolor[HTML]{cfe2f3}214 \\
&mammal &778 &\cellcolor[HTML]{fff2cc}633 &\cellcolor[HTML]{d9ead3}619 &\cellcolor[HTML]{cfe2f3}434 \\
&hibian &211 &\cellcolor[HTML]{fff2cc}148 &\cellcolor[HTML]{d9ead3}139 &\cellcolor[HTML]{cfe2f3}124 \\
&insect &366 &\cellcolor[HTML]{fff2cc}179 &\cellcolor[HTML]{d9ead3}176 &\cellcolor[HTML]{cfe2f3}145 \\
&fish &1089 &\cellcolor[HTML]{fff2cc}1054 &\cellcolor[HTML]{d9ead3}987 &\cellcolor[HTML]{cfe2f3}722 \\
&bird &739 &\cellcolor[HTML]{fff2cc}546 &\cellcolor[HTML]{d9ead3}545 &\cellcolor[HTML]{cfe2f3}480 \\
&reptile &279 &\cellcolor[HTML]{fff2cc}232 &\cellcolor[HTML]{d9ead3}231 &\cellcolor[HTML]{cfe2f3}210 \\
&celebrity &1514 &\cellcolor[HTML]{fff2cc}1484 &\cellcolor[HTML]{d9ead3}1466 &\cellcolor[HTML]{cfe2f3}732 \\
&instrument &477 &\cellcolor[HTML]{fff2cc}375 &\cellcolor[HTML]{d9ead3}368 &\cellcolor[HTML]{cfe2f3}277 \\
&plant &606 &\cellcolor[HTML]{fff2cc}601 &\cellcolor[HTML]{d9ead3}593 &\cellcolor[HTML]{cfe2f3}489 \\
&electronics &432 &\cellcolor[HTML]{fff2cc}354 &\cellcolor[HTML]{d9ead3}342 &\cellcolor[HTML]{cfe2f3}269 \\
&tool &801 &\cellcolor[HTML]{fff2cc}213 &\cellcolor[HTML]{d9ead3}209 &\cellcolor[HTML]{cfe2f3}150 \\
&transportation &334 &\cellcolor[HTML]{fff2cc}296 &\cellcolor[HTML]{d9ead3}290 &\cellcolor[HTML]{cfe2f3}227 \\
&sport &694 &\cellcolor[HTML]{fff2cc}478 &\cellcolor[HTML]{d9ead3}464 &\cellcolor[HTML]{cfe2f3}395 \\
&book &1030 &\cellcolor[HTML]{fff2cc}826 &\cellcolor[HTML]{d9ead3}777 &\cellcolor[HTML]{cfe2f3}645 \\
&household &475 &\cellcolor[HTML]{fff2cc}319 &\cellcolor[HTML]{d9ead3}299 &\cellcolor[HTML]{cfe2f3}221 \\
&car &500 &\cellcolor[HTML]{fff2cc}320 &\cellcolor[HTML]{d9ead3}320 &\cellcolor[HTML]{cfe2f3}208 \\
\cellcolor[HTML]{f4cccc}\textbf{Summary} &\cellcolor[HTML]{f4cccc}22 &\cellcolor[HTML]{f4cccc}14910 &\cellcolor[HTML]{f4cccc}10453 &\cellcolor[HTML]{f4cccc}10102 &\cellcolor[HTML]{f4cccc}7568 \\
\bottomrule
\end{tabular}
\end{adjustbox}
\label{table:more-filtering}
\end{table*}

\begin{figure}[htp]
  \centering
  \includegraphics[width=0.99\linewidth]{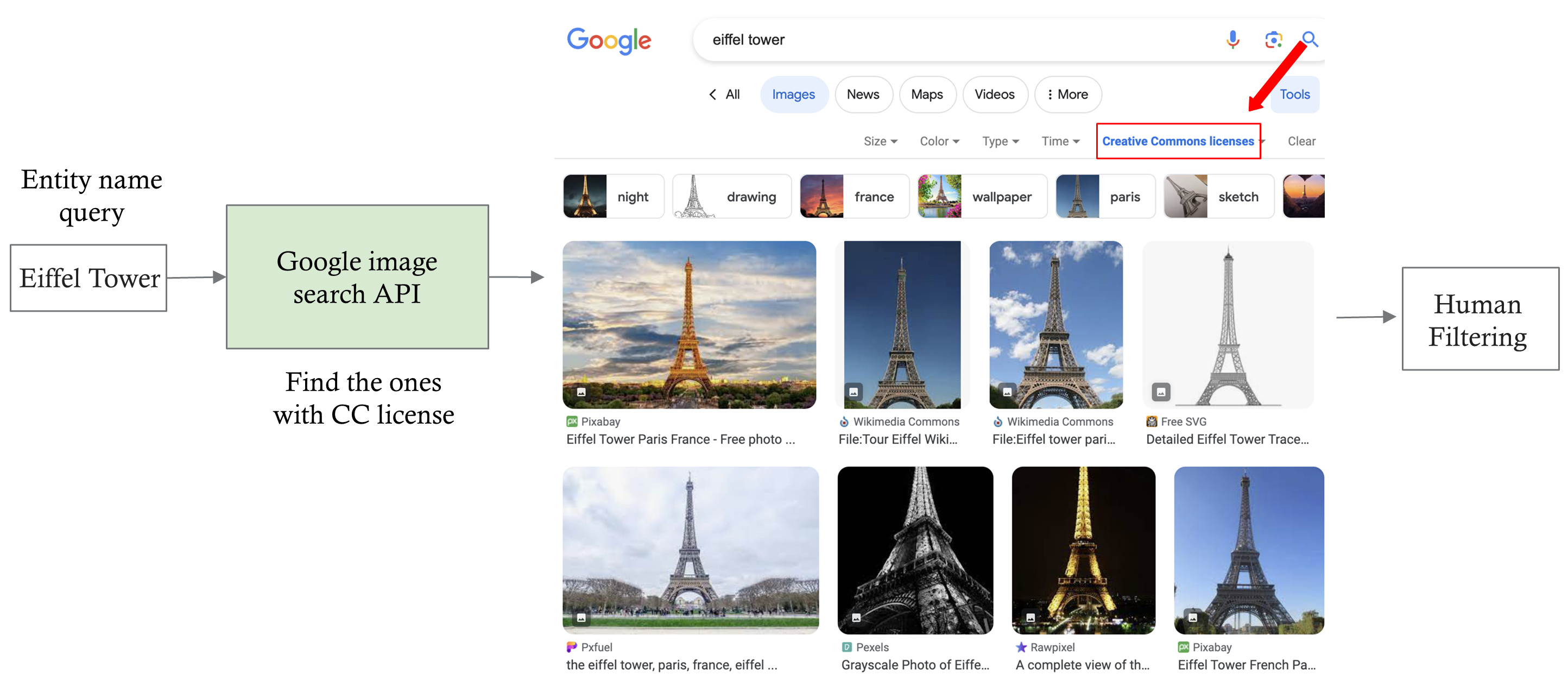}
  \caption{Collecting images for building the evaluation dataset. Licenses: CC Publicdomain, CC Attribute, AA Sharealike, CC Noncommercial, or CC Nonderived licenses. Metadata: image URLs, source page URLs, renamed image names, and the corresponding Wikipedia page URL.}
  \label{Fig:google_image}
\end{figure}

\section{Types of Questions}

More introduction of different types of question in the SnapNTell dataset are shown Table~\ref{table:appendix-type-questions}.

\begin{table*}[h]\small
\caption{Types of questions.}
\centering
\begin{adjustbox}{width=0.99\linewidth}
\begin{tabular}{p{3cm}|p{11cm}}
\toprule
Types of questions & Definition \\
\midrule
Static facts (absolute facts, discrete facts) & These are objective facts that are concrete and are not contingent on other conditions. They can usually be answered with a short, unique answer. For example:  When was Barack Obama born?
\\
\midrule
Narrative facts & These facts encompass comprehension of larger contexts (e.g., song lyrics, movie plot, historical events). They are factual in the sense that the content of the narrative should accurately reflect the source material or events, but a correct answer is usually not unique, as they can vary in their level of detail and focus. For example:  What is the plot of ``The Godfather”?
 \\
\midrule
Dynamic facts & These are facts that are subject to change over time. For example:  What is the Yelp customer rating of the Eleven Madison Park restaurant in NYC?
 \\
\midrule
Procedural facts & These are usually answers to “how” questions, outlining a sequence of steps to accomplish a task. While the steps may not be unique and could be subjective, in many cases, an answer can still be classified as logical (factual) or nonsensical (a hallucination). Note that these facts can overlap with dynamic facts or narrative facts. For example, How do you check the battery level of my Ray-Ban Stories Glasses?
 \\
\midrule
Subjective facts (opinion-based facts) & These “facts” are not objective, indisputable facts, but are based on individual perspectives or experiences. Recommendations fall in this category. While there’s generally no single correct answer to questions seeking subjective facts, it still requires the system to understand the topic and provide reasonable answers grounded by world facts. For example:  Where should I visit Tokyo next month?
\\
\bottomrule
\end{tabular}
\label{table:appendix-type-questions}
\end{adjustbox}
\vspace{10pt}
\end{table*}

\section{Method}\label{sec:method-appendix}

\begin{figure}[htp]
  \centering
  \includegraphics[width=0.99\linewidth]{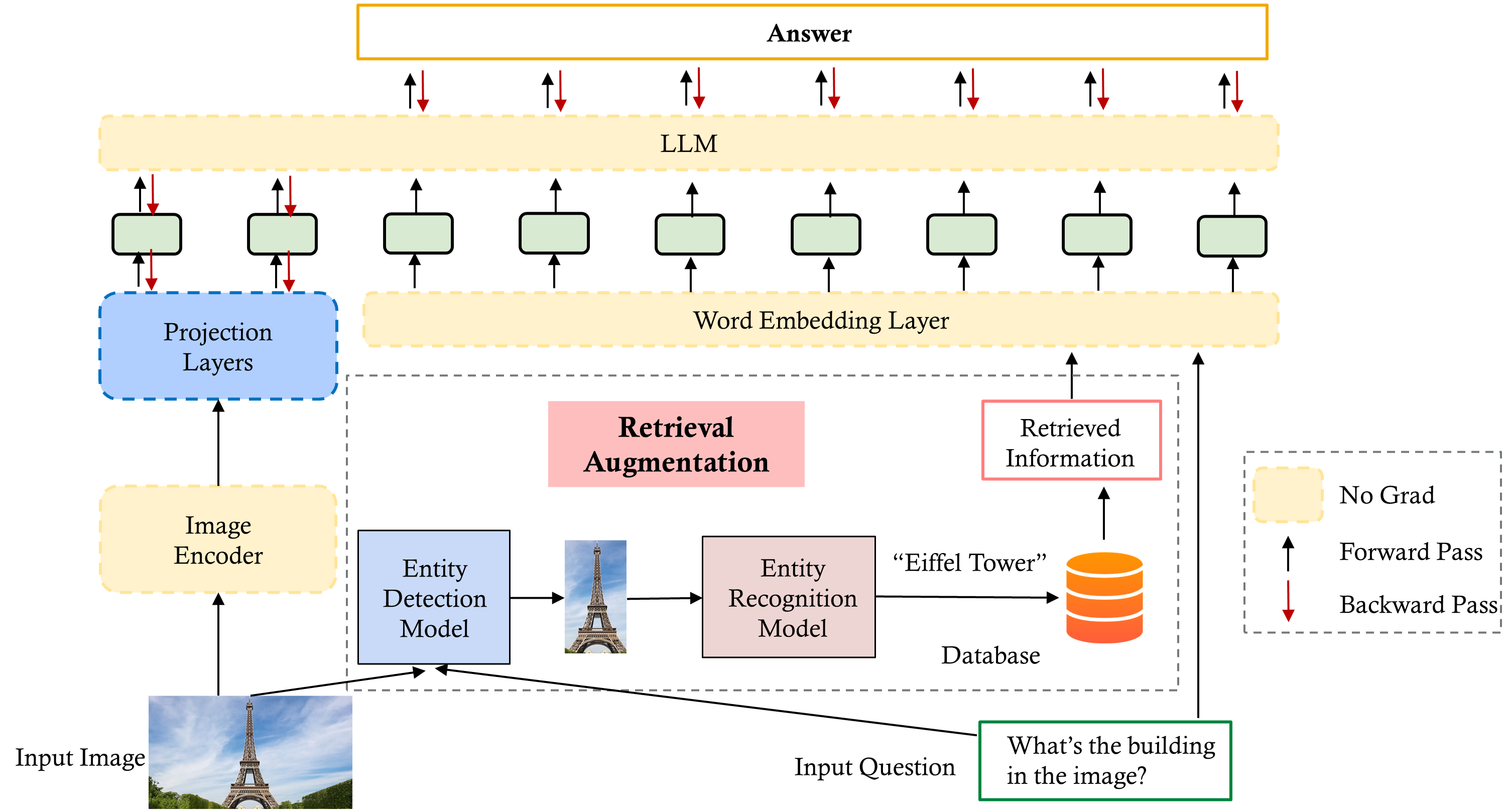}
  \caption{The architecture of our SnapNTell model. The input to the model is an image-question pair, and our model first uses retrieval augmentation to retrieve useful information regarding the entity in the image. Then, the retrieved information is combined with the question as input to the word embedding layer, where the text embeddings will be combined with image-projected embeddings as the input to LLM, which finally generates a knowledgeable answer as the output. }
  \label{Fig:propsoed_approach}
\end{figure}

In this section, we will introduce the details of our proposed retrieval-augmented multimodal LLM model. The architecture of our model is shown in Figure~\ref{Fig:propsoed_approach}. Our model can be considered twofold: (1) \textbf{Retrieval augmentation}. Given the input image-question pair, we retrieve useful entity-centric information within knowledge sources. (2) \textbf{Entity-centric knowledge-based answer generation}. The retrieved information will be combined with the image and question together to generate the answer. More details are introduced in the following sections.

\subsection{Retrieval Augmentation}

The retrieval augmentation process can be subdivided into three distinct steps: (i) Semantic region extraction via language-guided object detection, (ii) Entity recognition via image retrieval, and (iii) Knowledge retrieval via multi-source aggregation.

\paragraph{Semantic Region Extraction via Language-Guided Object Detection}
Due to the presence of entities within the image that occupy only a portion of the available space, employing a comprehensive image-level entity recognition approach may lead to a decrease in recognition performance. Instead, we opt to initially extract the image region containing the entity and utilize this specific region in subsequent recognition processes to enhance accuracy.
During this phase, we leverage a language-guided object detection model, i.e., GLIP \citep{Li2021GroundedLP}, to extract meaningful regions from complex images. This approach helps precisely identify and extract image regions directly relevant to specific textual queries. It accomplishes this by understanding the context of the query and adjusting its object detection method to find the most important image areas. This step enables the system to better understand the query's context, resulting in more accurate and contextually meaningful region extraction.

\paragraph{Entity Recognition via Image Retrieval}
To accomplish this goal, we begin by constructing a similarity index using CLIP embeddings, specifically employing Faiss \citep{Johnson2017BillionScaleSS} as our indexing tool. Our indexing database is established based on the WIT dataset \citep{Srinivasan2021WITWI}. This database follows a key-value mapping structure, where the keys represent CLIP ViT-B/32 image embeddings, and the corresponding text descriptions serve as the values. Faiss, known for its efficiency in similarity search, is utilized for indexing \citep{Johnson2017BillionScaleSS}.

Once the indexing database is set up, we are ready to proceed with the query process. Given an input query image, denoted as $I$ (which is the entity image region extracted in the preceding step), we perform a $k$-nearest neighbor retrieval based on cosine similarity between the embeddings of the query image and those of the database images. The retrieval outcomes are represented as $\mathcal{R}(I)=\{\left(i_1, c_1\right), \cdots,\left(i_k, c_k\right)\}$, where for each $j$ within the range of $1$ to $k$, $i_j$ and $c_j$ correspond to the retrieved image and its associated caption, respectively.
Subsequently, by using the extracted image region as input for a search in the indexing database, we identify the entity within the extracted image region. This identification is achieved by comparing it with the most similar images retrieved from the indexing database, ultimately resulting in image-level entity recognition.

\paragraph{Knowledge Retrieval via Multi-Source Aggregation}

Given the wide array of questions users may pose, we need to obtain additional information to compile the necessary resources for crafting accurate responses. Furthermore, certain queries may demand the latest information, which is not readily available within pre-existing databases or knowledge graphs. In such cases, we rely on external sources of knowledge, such as online references, to gather essential data, encompassing elements like ``year built," ``description," and other pertinent details.
To accomplish this, we leverage Knowledge Graph (KG) and conduct web searches to access relevant knowledge connections. This approach enables us to acquire a wealth of information concerning the specified image region, thereby bolstering our capacity to grasp and contextualize the extracted content effectively.

\subsection{Entity-centric Knowledge-based Answer Generation}

Following the preceding step, where we've gathered insightful information from diverse sources, we now proceed to the second phase: determining how to integrate the input image, the question, and the retrieved information in order to produce a knowledge-driven response.

Our approach is illustrated in Figure~\ref{Fig:propsoed_approach}.
Our strategy for improving the model's multimodal comprehension entails pre-training a LLM using paired multimodal data, which comprises images alongside corresponding textual descriptions. To achieve this, we draw inspiration from \citet{Moon2023AnyMALAE} and create lightweight adapters for each modality. These adapters facilitate the transformation of inputs into the text token embedding space of a designated LLM.

Our approach transforms the text token embedding space of the LLM into a unified token embedding space, where tokens can represent either textual or image content. The number of token embeddings allocated to each input modality is predetermined for each adapter, ranging from 64 to 256. 
Throughout the alignment training process, we keep the model parameters of the underlying LLM frozen. This approach not only accelerates convergence compared to training the model from scratch but also allows the model to inherit the reasoning capabilities of the LLM during inference.
Additionally, to maximize feature compatibility, we employ an encoder denoted as $g(\cdot)$ for the image modality. This encoder has previously been aligned with a text embedding space, for instance, in the case of CLIP \citep{Radford2021LearningTV, Schuhmann2022LAION5BAO}. For each pair of text and image, represented as $(\mathbf{X}_{\texttt{text}}, \mathbf{X}_{\texttt{image}})$, we align them using specific objectives along with a projection module, such as the Perceiver Resampler \citep{Alayrac2022FlamingoAV} for the vision encoder. 

\begin{equation}
\begin{aligned}
p(\mathbf{X}_{\texttt{text}} | \mathbf{X}_{\texttt{image}}) = \prod_{i=1}^{L} p_{\theta}(\mathbf{X}^{[i]}_{\texttt{text}} | \mathbf{Z}_{\texttt{image}}, \mathbf{Z}^{[1:i\small{-}1]}_{\texttt{text}}) 
\end{aligned}
\end{equation}
\begin{equation}
\begin{aligned}
\mathbf{Z}_{\texttt{image}} = \texttt{Proj}_{\theta}(h_{\texttt{latents}}, g(\mathbf{X}_{\texttt{image}}))
\end{aligned}
\end{equation}

\section{More Related Works}\label{sec:related-work-appendix}

\paragraph{Knowledge-based VQA}

Various vision-language tasks often require knowledge to answer questions based on image content and have evolved in recent years. Beginning with datasets like FVQA \citep{Wang2016FVQAFV}, which extracted facts from pre-established knowledge bases, the field has progressed to more challenging ones like the OK-VQA dataset \citep{Marino2019OKVQAAV}, encompassing diverse knowledge categories. MultiModalQA  \citep{Talmor2021MultiModalQACQ} introduced complexity with questions demanding cross-modal reasoning over snippets, tables, and images. The successor of OK-VQA, AOK-VQA \citep{Schwenk2022AOKVQAAB}, raises the bar by providing questions that transcend simple knowledge base queries. ManyModalQA \citep{Hannan2020ManyModalQAMD} shifts the focus to answer modality selection, MIMOQA \citep{Singh2021MIMOQAMI} emphasizes multimodal answer extraction, and WebQA \citep{Chang2021WebQAMA} introduces real-world knowledge-seeking questions, albeit with some limitations regarding entity categorization and granularity. More comparison details are introduced in Section~\ref{sec:compare_datasets}.

\paragraph{Multimodal LLMs} 

Expanding text-only LLMs to interpret visual information typically involves integrating a visual encoder with a frozen LLM, using extensive image captioning datasets for alignment \citep{Koh2023GroundingLM,Wu2023NExTGPTAM,Chowdhery2022PaLMSL}. This integration can be accomplished through methods such as adapter-based tuning \citep{Alayrac2022FlamingoAV}, which fine-tunes a small portion of the model to process visual inputs, or prefix tuning \citep{Tsimpoukelli2021MultimodalFL}, where trained prefixed vectors are inputted to guide the frozen LLM towards contextually relevant text outputs based on the visual data. These techniques allow LLMs to maintain their linguistic prowess while gaining visual understanding without full model retraining \citep{Yin2023ASO}.

\paragraph{Retrieval augmented LLM} 
Several prior approaches have investigated retrieval-augmented in the text-only setting or image captioning tasks. 
\citet{Guu2020REALMRL} augmented language model pretraining with a latent knowledge retriever, which allows the model to retrieve and attend over documents from a large corpus such as Wikipedia, used during pretraining, fine-tuning, and inference. 
\citet{Srinivasan2022QUILLQI} demonstrated that retrieval augmentation of queries provides LLMs with valuable additional context, enabling improved understanding. 
\citet{Yasunaga2023RetrievalAugmentedML} proposed a retriever to retrieve relevant multimodal documents from external memory and use the generator to make predictions for the input.
\citet{Yang2023InferenceWR} proposed an accelerator to losslessly speed up LLM inference with references through retrieval.
\citet{Yang2023ReViLMRV} introduced a retrieval-augmented visual language model, built upon the Flamingo \citep{Alayrac2022FlamingoAV}, which supports retrieving the relevant knowledge from the external database for zero and in-context few-shot image captioning. 
Another related work by \citet{Gui2021KATAK} integrated implicit and explicit knowledge in an encoder-decoder architecture for jointly reasoning over both knowledge sources during answer generation.

\paragraph{Open-domain visual entity recognition}   

\citet{Hu2023OpendomainVE} introduced Open-domain Visual Entity Recognition (OVEN) for linking images to Wikipedia entities through text queries. \citet{Chen2023CanPV} presented INFOSEEK, a Visual Question Answering dataset designed for information-seeking queries. OVEN excels at entity recognition but relies on a knowledge base for entity names, while INFOSEEK primarily provides factual answers. Our research aims to bridge these gaps by generating informative paragraphs that offer context, enabling a deeper understanding beyond mere facts.

\section{More Statistics of the SnapNTell Dataset}\label{sec:appendix-more-statistics}

\begin{figure}[htp]
  \centering
  \includegraphics[width=0.8\linewidth]{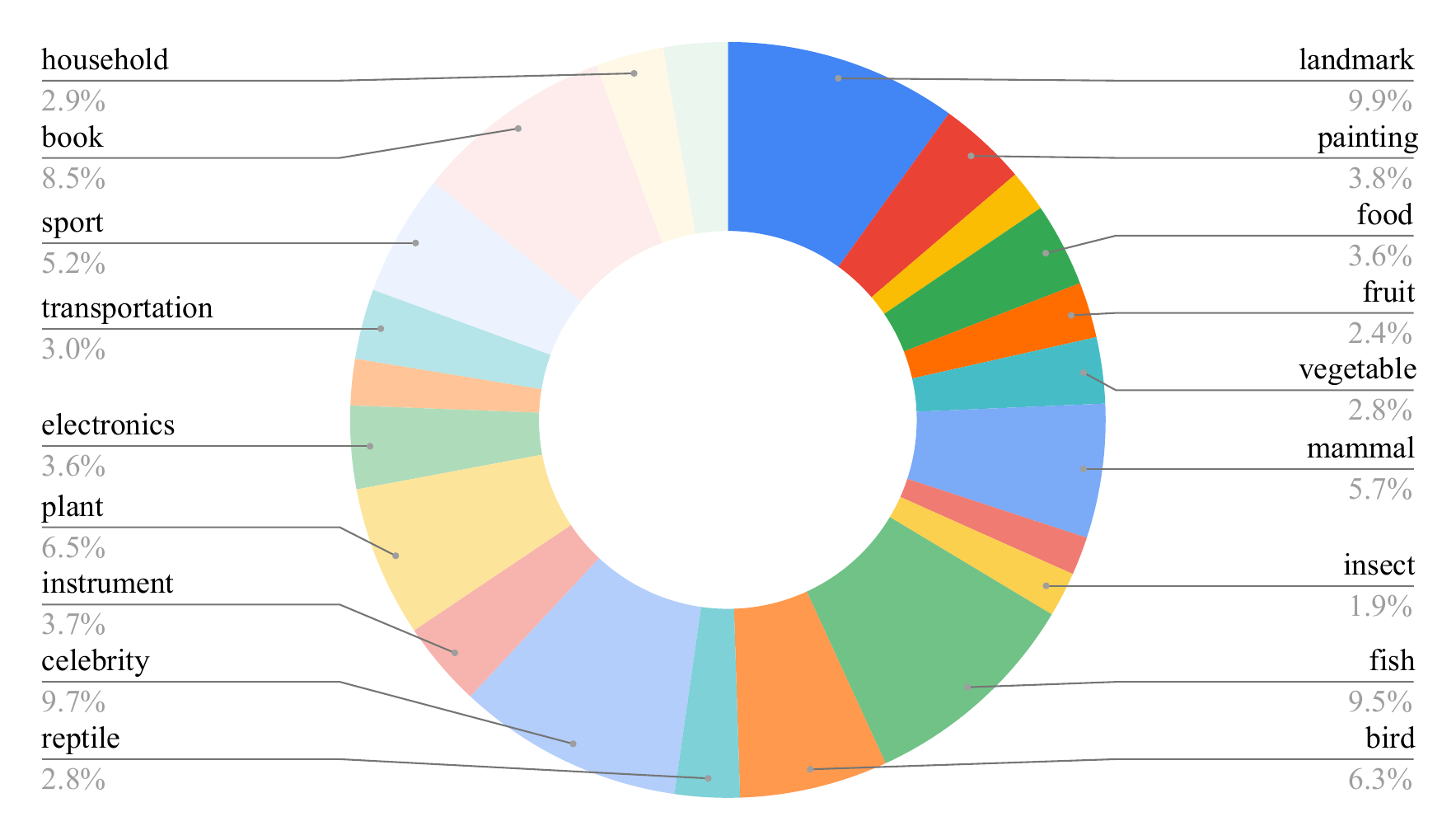}
  \caption{Statistics of number of entities in each category.}
  \label{Fig:Statistics_catergory}
  \vspace{-10pt}
\end{figure}

\begin{figure}[htp]
  \centering
  \includegraphics[width=0.8\linewidth]{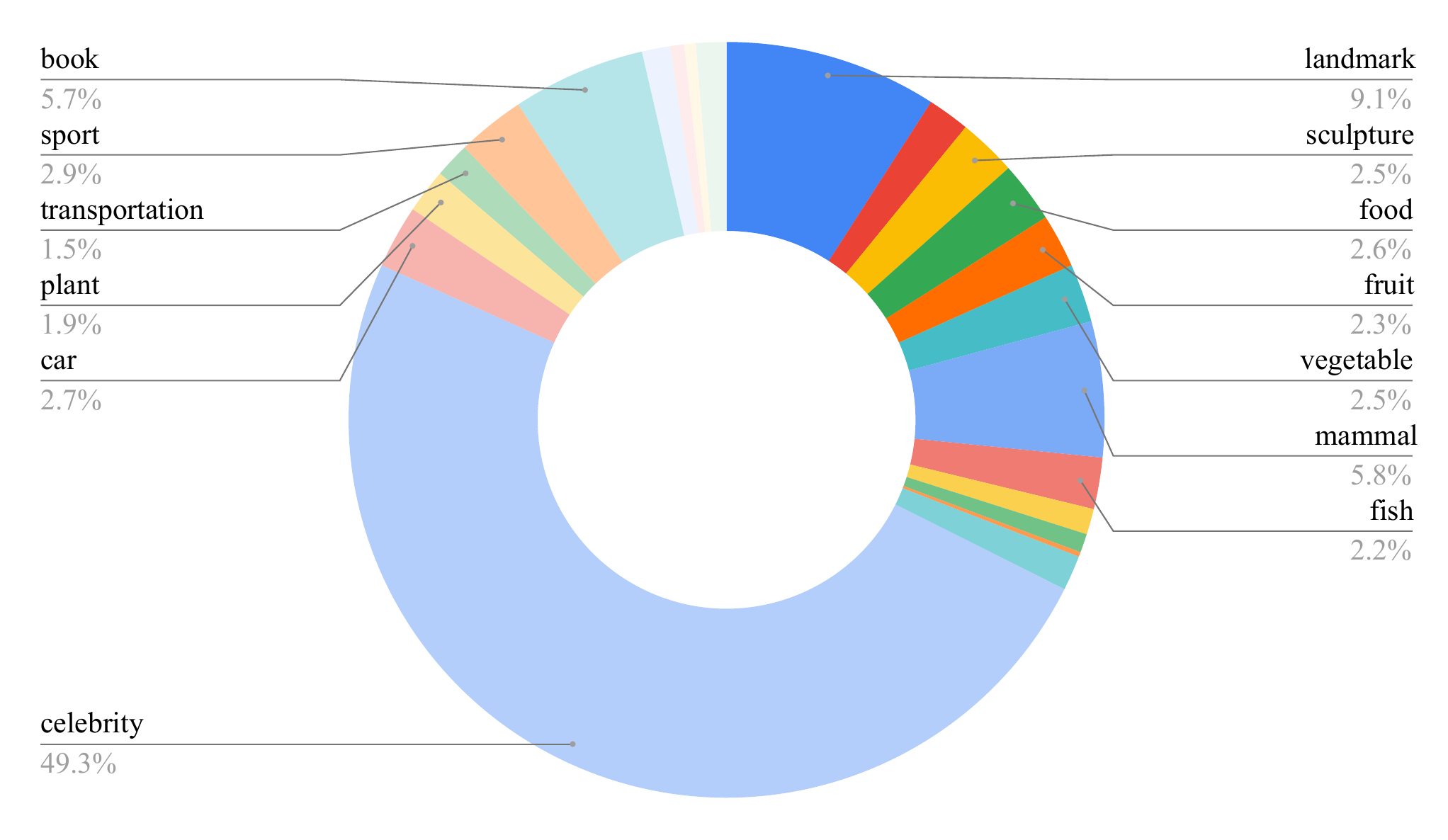}
  \caption{Statistics of all pageviews for all categories.}
  \label{Fig:statistics_pageview}
  \vspace{-10pt}
\end{figure}

\begin{figure}[htp]
  \centering
  \includegraphics[width=0.8\linewidth]{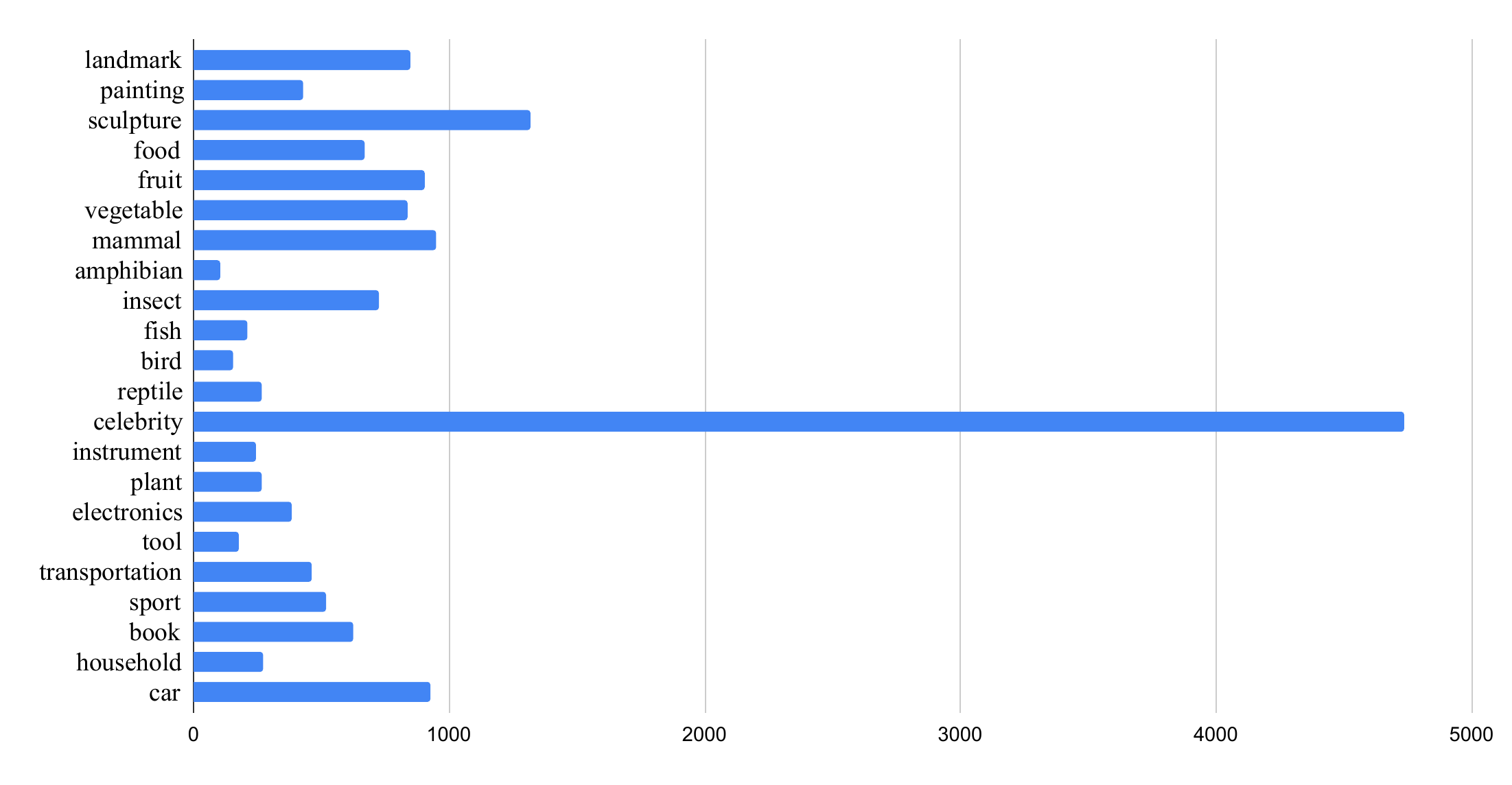}
  \caption{Average pageview per entity within each category, where average pageview is defined as the sum of pageviews/ number of entity.}
  \label{Fig:ave_pageview}
  \vspace{-10pt}
\end{figure}

In Table~\ref{table:dataset-statistics} and Figure~\ref{Fig:Statistics_catergory},\ref{Fig:statistics_pageview},\ref{Fig:ave_pageview}, we show more statistics of the SnapNTell dataset.

\begin{figure*}[htp]
  \centering
  \includegraphics[width=0.9\linewidth]{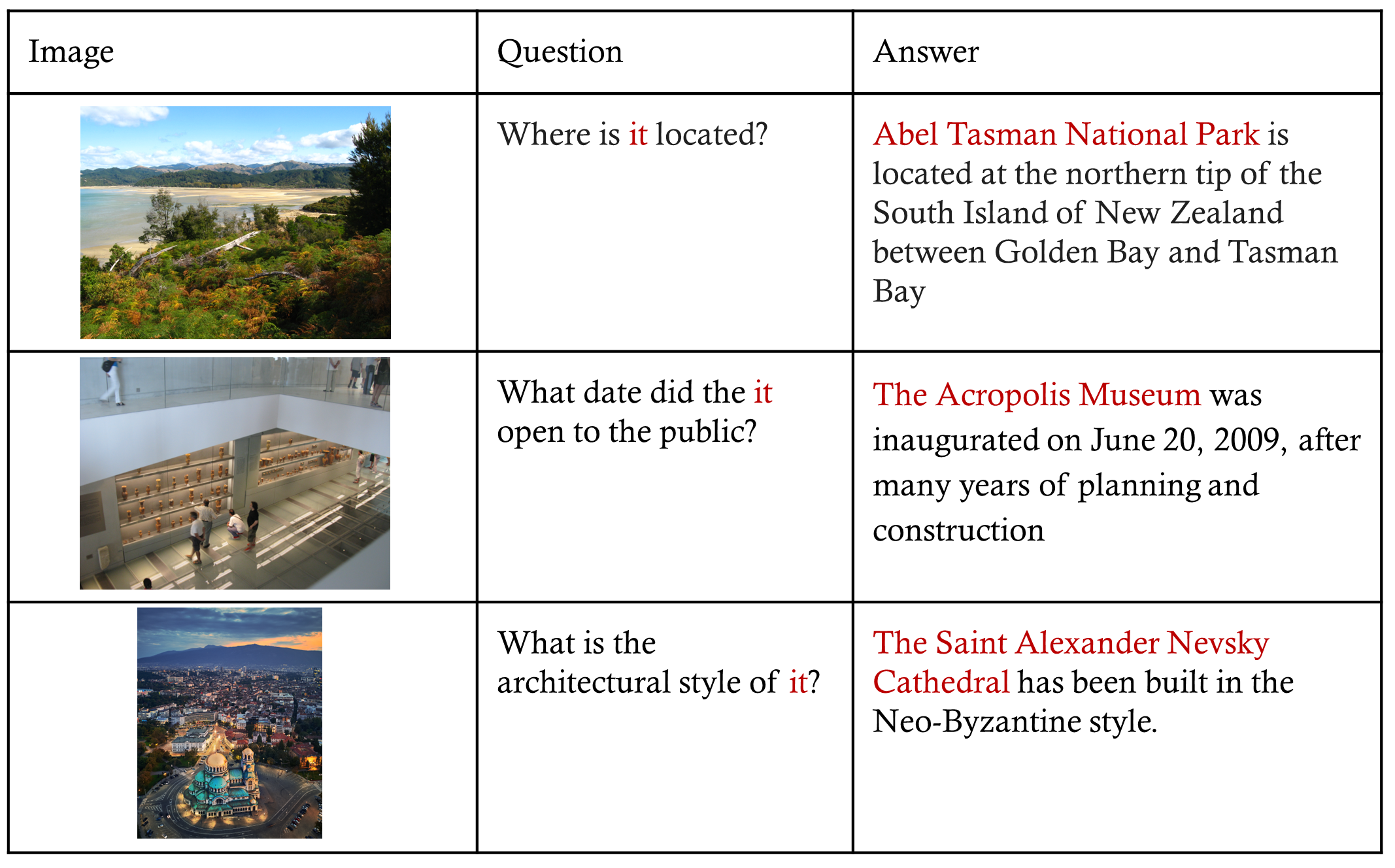}
  \caption{Examples from our SnapNTell dataset.}
  \label{Fig:our_example}
\end{figure*}

\begin{table}[htp]\centering
\caption{Category statistics of the SnapNTell dataset.}
\vspace{-5pt}
\begin{adjustbox}{width=0.4\linewidth}
\begin{tabular}{lcc}
\toprule
&\textbf{Category} &\textbf{Number of entities} \\
\midrule
\multirow{22}{*}{\textbf{Category}} &landmark &753 \\
&painting &288 \\
&sculpture &134 \\
&food &271 \\
&fruit &180 \\
&vegetable &214 \\
&mammal &434 \\
&fish &124 \\
&bird &145 \\
&reptile &722 \\
&amphibian &480 \\
&insect &210 \\
&celebrity &732 \\
&instrument &277 \\
&plant &489 \\
&electronics &269 \\
&tool &150 \\
&transportation &227 \\
&sport &395 \\
&book &645 \\
&household &221 \\
&car &208 \\
\midrule
\cellcolor[HTML]{ffff00}\textbf{Summary} &\cellcolor[HTML]{ffff00}\textbf{22} &\cellcolor[HTML]{ffff00}\textbf{7568} \\
\bottomrule
\end{tabular}
\end{adjustbox}
\label{table:dataset-statistics}
\vspace{-5pt}
\end{table}

\clearpage

\section{Some Result Examples and Human Evaluation}


In Table~\ref{table:human}, we showed several example result by different models, and the corresponding human evaluation results.

\begin{table*}[!ht]
\centering
\caption{Examples of answers generated by different models, where Ground-truth, BLIP2, MiniGPT4, Open-Flamingo, InstructBLIP, LLaVA, SnapNTell are assigned as M0, M1, ..., M6 in rating.}
\begin{adjustbox}{width=0.85\linewidth}
\begin{tabular}{p{3.5cm}p{12cm}}
\toprule
\textbf{Image} &
\begin{minipage}{.6\textwidth}
\includegraphics[height=40mm]{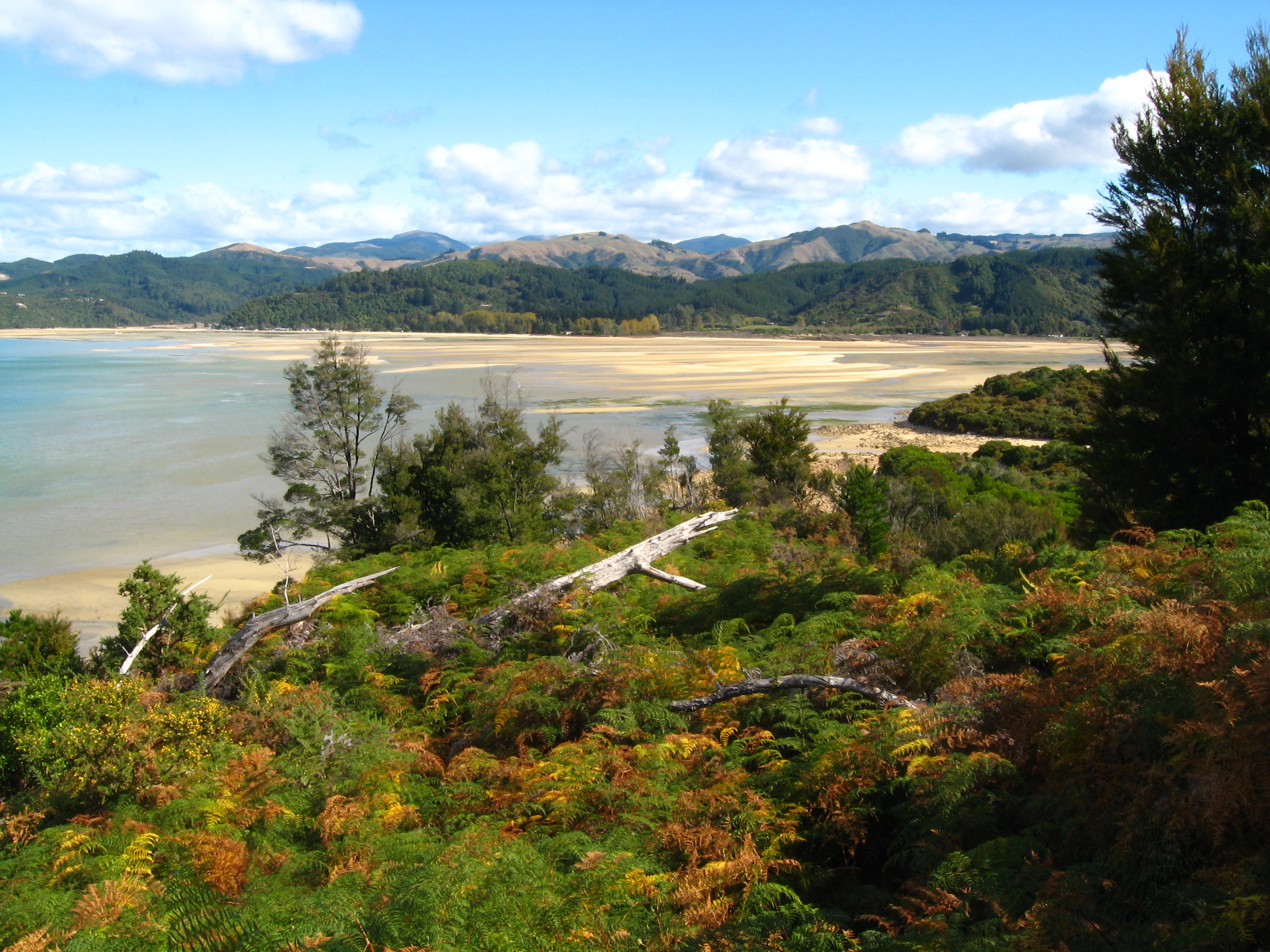}
\vspace{2pt} \end{minipage}
\\
\textbf{Question} & \textbf{\textit{Where is the attraction located?}} \\
\midrule 
Ground-truth (M0) &Abel Tasman National Park is located at the northern tip of the South Island of New Zealand between Golden Bay and Tasman Bay   \\ \cmidrule(r){2-2}
BLIP-2 (M1) &california   \\ \cmidrule(r){2-2}
Mini-GPT4 (M2) &The image shows a beach with a body of water in the foreground and trees and mountains in the background.  \\ \cmidrule(r){2-2}    
Open-Flamingo (M3) &It is located in New Zealand.  \\ \cmidrule(r){2-2} 
Instruct-BLIP (M4) &beach 1  \\ \cmidrule(r){2-2}   
LLaVA (M5) &The image is located at the beach, which is surrounded by a forest of trees.  \\ \cmidrule(r){2-2}   
SnapNTell (Ours) (M6) &The image is not located in the Republic of Altai, but rather in the South Island of New Zealand. The location is the Abel Tasman National Park, specifically the area around Awaroa Beach. \\ \cmidrule(r){2-2}   
\midrule
Human Rating &$\text{M}0=\text{M}6>\text{M}3>\text{M}1=\text{M}2=\text{M}5>\text{M}4$ \\
\bottomrule
\end{tabular}
\end{adjustbox}
\label{table:human}
\end{table*}

\begin{table*}[!ht]
\centering
\begin{adjustbox}{width=0.85\linewidth}
\begin{tabular}{p{3.5cm}p{12cm}}
\toprule
\textbf{Image} &
\begin{minipage}{.6\textwidth}
\includegraphics[height=40mm]{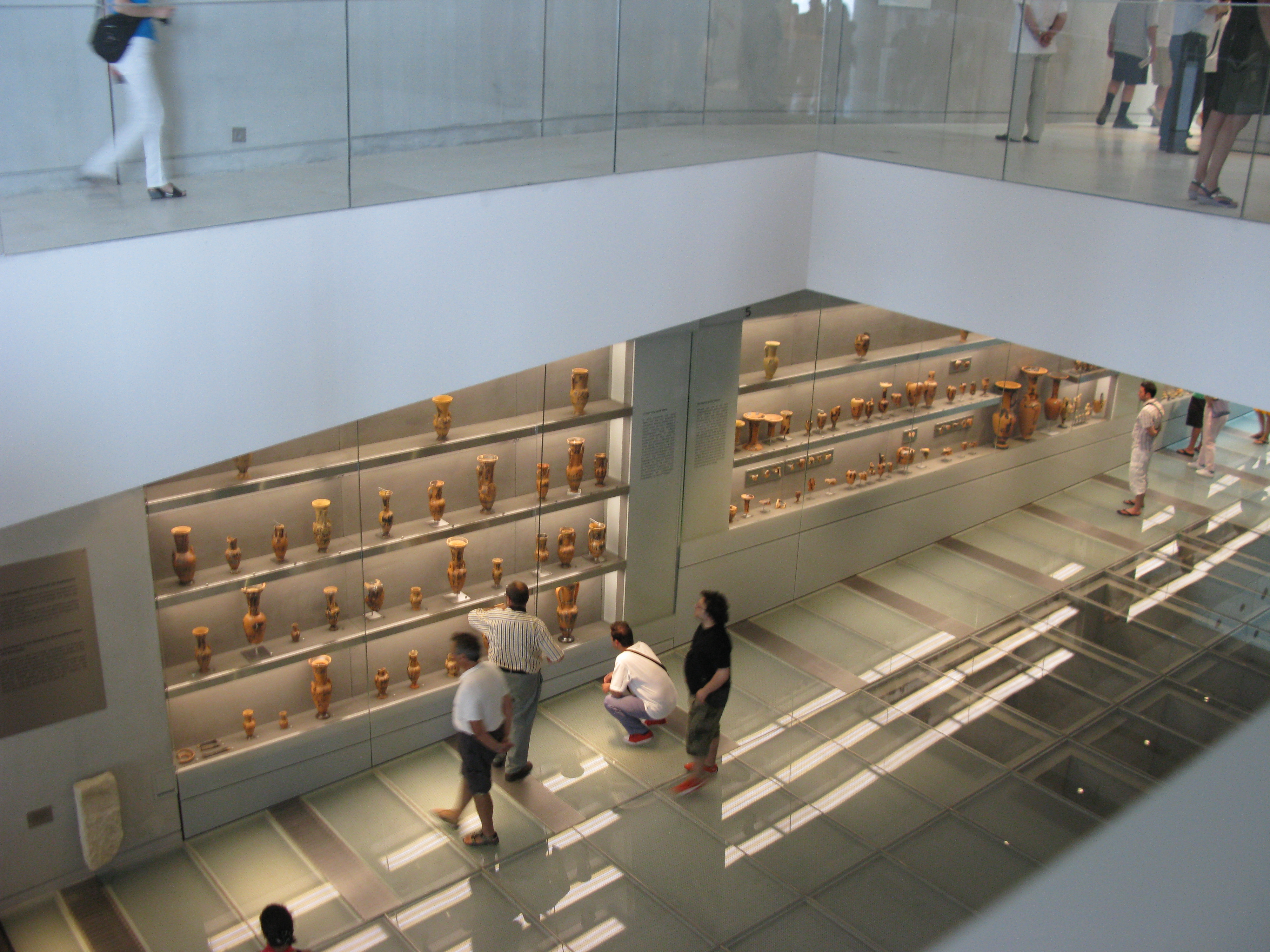}
\vspace{2pt} \end{minipage}
\\
\textbf{Question} & \textbf{\textit{What date did it open to the public?}} \\
\midrule 
Ground-truth (M0) &The Acropolis Museum was inaugurated on June 20, 2009 after many years of planning and construction   \\ \cmidrule(r){2-2}
BLIP-2 (M1) &1 January 2009   \\ \cmidrule(r){2-2}
Mini-GPT4 (M2) &The museum opened to the public on January 1, 2018 \\ \cmidrule(r){2-2}    
Open-Flamingo (M3) &The museum opened to the public on December 12, 1987.  \\ \cmidrule(r){2-2} 
Instruct-BLIP (M4) &2007  \\ \cmidrule(r){2-2}   
LLaVA (M5) &The museum opened to the public on April 13, 2019.   \\ \cmidrule(r){2-2}   
SnapNTell (Ours) (M6) &The Acropolis Museum opened to the public on June 20, 2009. \\ \cmidrule(r){2-2}   
\midrule
Human Rating &$\text{M}0=\text{M}6>\text{M}1>\text{M}2=\text{M}3=\text{M}4=\text{M}5$ \\
\bottomrule
\end{tabular}
\label{table:appendix-result-example}
\end{adjustbox}
\end{table*}

\begin{table*}[!ht]
\centering
\begin{adjustbox}{width=0.85\linewidth}
\begin{tabular}{p{3.5cm}p{12cm}}
\toprule
\textbf{Image} &
\begin{minipage}{.6\textwidth}
\includegraphics[height=40mm]{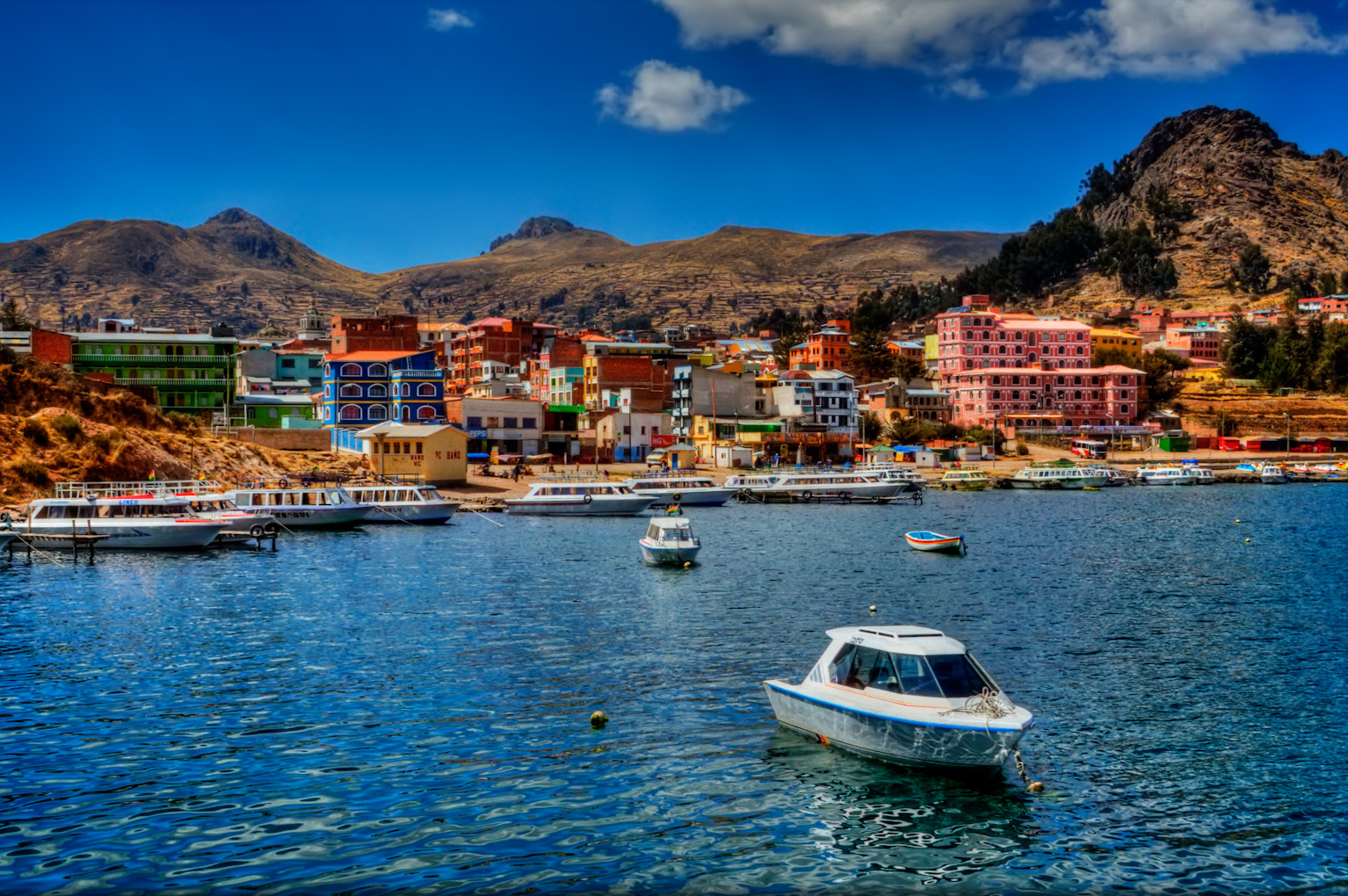}
\vspace{2pt} \end{minipage}
\\
\textbf{Question} & \textbf{\textit{What is the name of the view in this picture?}} \\
\midrule 
Ground-truth (M0) &Lake Titicaca is a large freshwater lake located in the Andes Mountains spanning the borders of both Bolivia and Peru   \\ \cmidrule(r){2-2}
BLIP-2 (M1) &a town in Peru   \\ \cmidrule(r){2-2}
Mini-GPT4 (M2) &It is a small town with colorful buildings and boats in the water. \\ \cmidrule(r){2-2}  
Open-Flamingo (M3) &This is the city of Puerto Natales in Chile. \\ \cmidrule(r){2-2} 
Instruct-BLIP (M4) &harbor 2   \\ \cmidrule(r){2-2}   
LLaVA (M5) &It is a cityscape with boats in the water, located near a mountainous area.   \\ \cmidrule(r){2-2}   
SnapNTell (Ours) (M6) &It's Lake Titicaca in the Andes Mountains. \\ \cmidrule(r){2-2}   
\midrule
Human Rating &$\text{M}0=\text{M}6>\text{M}1>\text{M}3>\text{M}2=\text{M}5>\text{M}4$ \\
\bottomrule
\end{tabular}
\end{adjustbox}
\end{table*}

\begin{table*}[!ht]
\centering
\begin{adjustbox}{width=0.85\linewidth}
\begin{tabular}{p{3.5cm}p{12cm}}
\toprule
\textbf{Image} &
\begin{minipage}{.6\textwidth}
\includegraphics[height=40mm]{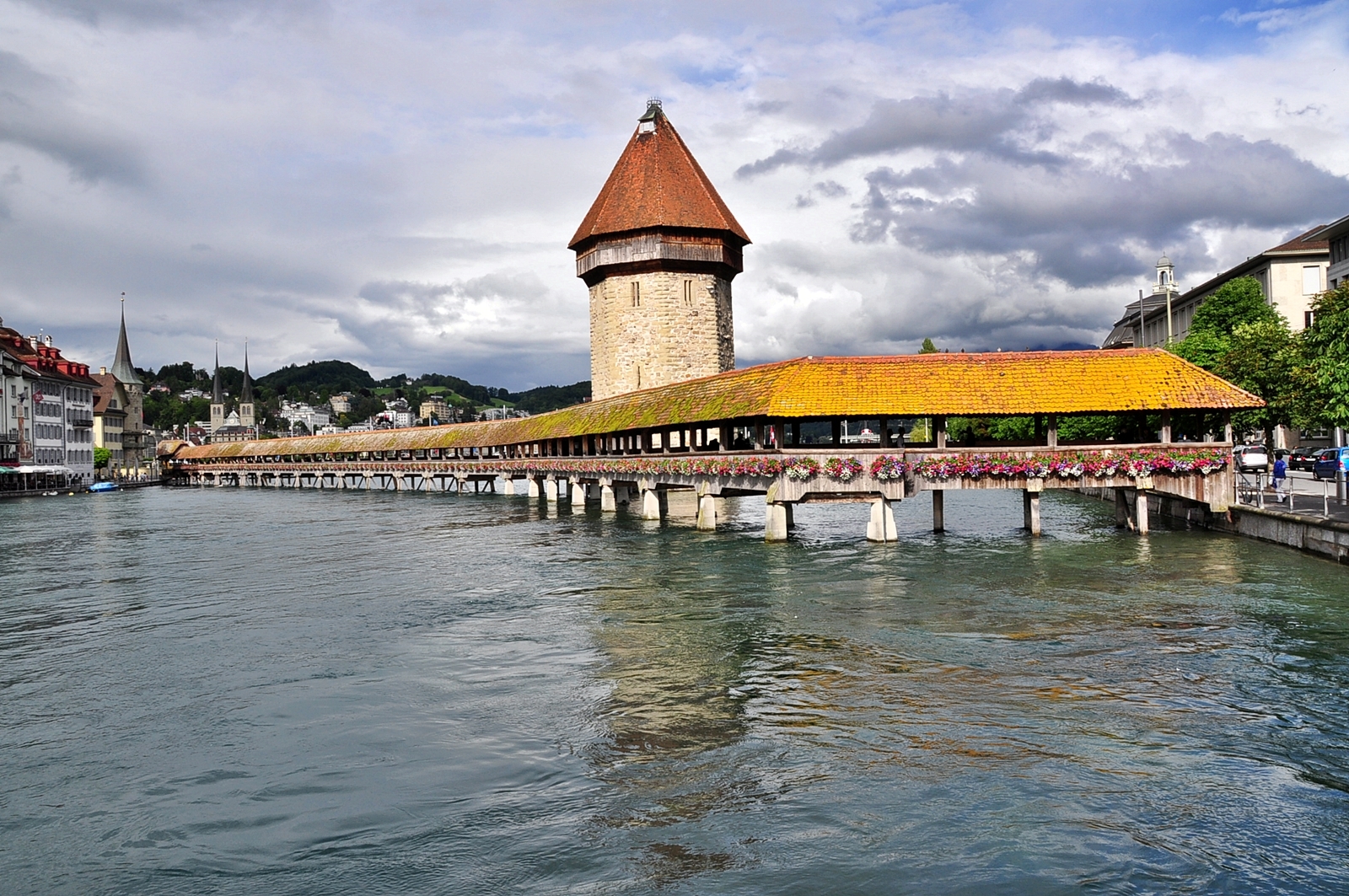}
\vspace{2pt} \end{minipage}
\\
\textbf{Question} & \textbf{\textit{What is the age of the building now?}} \\
\midrule 
Ground-truth (M0) & The Chapel Bridge is over 700 years old having been built in 1333  \\ \cmidrule(r){2-2}
BLIP-2 (M1) & 16th century  \\ \cmidrule(r){2-2}
Mini-GPT4 (M2) & The age of the building in the image is not known. \\ \cmidrule(r){2-2}  
Open-Flamingo (M3) &The age of the bridge is over 100 years old.  \\ \cmidrule(r){2-2} 
Instruct-BLIP (M4)  & old \\ \cmidrule(r){2-2}   
LLaVA (M5) &The age of the building in the image is not specified. However, it is described as an old building, which suggests that it could be quite old or historic.  \\ \cmidrule(r){2-2}   
SnapNTell (Ours) (M6) & The Chapel Bridge (Chapel Bridge or Kapellbrücke) in Lucerne, Switzerland, is a medieval wooden bridge that dates back to the 14th century. It was built in 1333 and is one of the oldest covered bridges in Europe, which is over 700 years old now.\\ \cmidrule(r){2-2}   
\midrule
Human Rating &$\text{M}6>=\text{M}0>\text{M}1=\text{M}2=\text{M}3=\text{M}4=\text{M}5$ \\
\bottomrule
\end{tabular}
\end{adjustbox}
\end{table*}

\end{document}